\documentclass[10pt,twocolumn,letterpaper]{article}

\usepackage{templates/cvpr}
\usepackage{times}
\usepackage{epsfig}
\usepackage{graphicx}
\usepackage{amsmath}
\usepackage{amssymb}
\usepackage{gensymb}
\usepackage{subcaption}
\usepackage[]{algorithm2e}
\usepackage{multirow}
\usepackage{enumitem}
\usepackage[square,sort,comma,numbers,compress]{natbib}
\usepackage{flushend}
\usepackage{placeins}

\usepackage[pagebackref=true,breaklinks=true,letterpaper=true,colorlinks,bookmarks=false]{hyperref}

\cvprfinalcopy 

\def\cvprPaperID{2150} 

\ifcvprfinal\fi
\begin{document}

\title{DeepNav: Learning to Navigate Large Cities}

\author{Samarth Brahmbhatt\\
Georgia Institute of Technology\\
Atlanta USA\\
{\tt\small samarth.robo@gatech.edu}
\and
James Hays\\
Georgia Institute of Technology\\
Atlanta USA\\
{\tt\small hays@gatech.edu}
}

\maketitle

\begin{abstract}
We present DeepNav, a Convolutional Neural Network (CNN) based algorithm for navigating large cities using locally visible street-view images. The DeepNav agent learns to reach its destination quickly by making the correct navigation decisions at intersections. We collect a large-scale dataset of street-view images organized in a graph where nodes are connected by roads. This dataset contains 10 city graphs and more than 1 million street-view images. We propose 3 supervised learning approaches for the navigation task and show how A* search in the city graph can be used to generate supervision for the learning. Our annotation process is fully automated using publicly available mapping services and requires no human input. We evaluate the proposed DeepNav models on 4 held-out cities for navigating to 5 different types of destinations. Our algorithms outperform previous work that uses hand-crafted features and Support Vector Regression (SVR)~\cite{mcdonalds}.
\end{abstract}

\section{Introduction}
Man-made environments like houses, buildings, neighborhoods and cities have a \textit{structure} - microwaves are found in kitchens, restrooms are usually situated in the corners of buildings, and restaurants are found in specific kinds of commercial areas. This structure is also \textit{shared} across environments - for example, most cities will have restaurants in their business district. It has been a long-standing goal of computer vision to learn this structure and use it to guide exploration of unknown man-made environments like unseen cities, buildings and houses. Knowledge of such structure can be used to recognize tougher visual concepts like occluded or small objects~\cite{3dgp, mottaghi}, to better delimit the boundaries of objects~\cite{struct_seg, cnn_crf}, and for robotic automation tasks like deciding drivable terrain for robots~\cite{struct_robotics}, etc.\par
In this paper, we address a task that requires understanding the large-scale structure of cities --  navigation in a new city to reach a destination in as few steps as possible. The agent neither has a map of the environment, nor does it know the location of the destination or itself. All it knows is that it needs to reach a particular type of destination, e.g. go to the nearest gas station in the city. The naive approach is a random walk of the environment. But if the agent has some learnt model of the structure of cities, then it can make \textit{informed decisions} - for example, gas stations are very likely to be found near freeway exits.\par
\begin{figure}[t]
\centering
  \begin{subfigure}[b]{0.23\textwidth}
    \includegraphics[width=\textwidth]{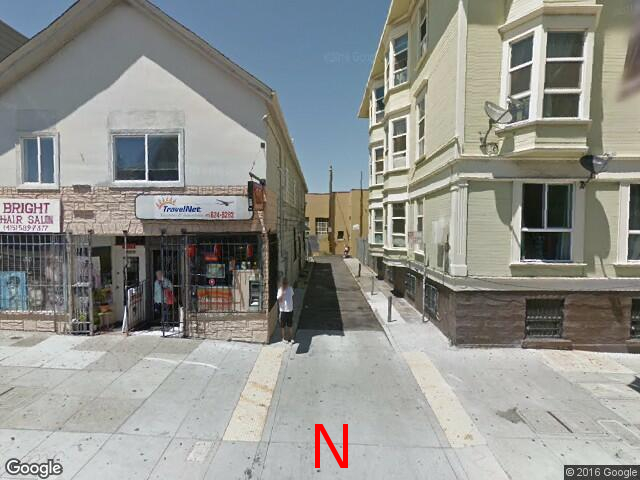}
  \end{subfigure}
  \begin{subfigure}[b]{0.23\textwidth}
    \includegraphics[width=\textwidth]{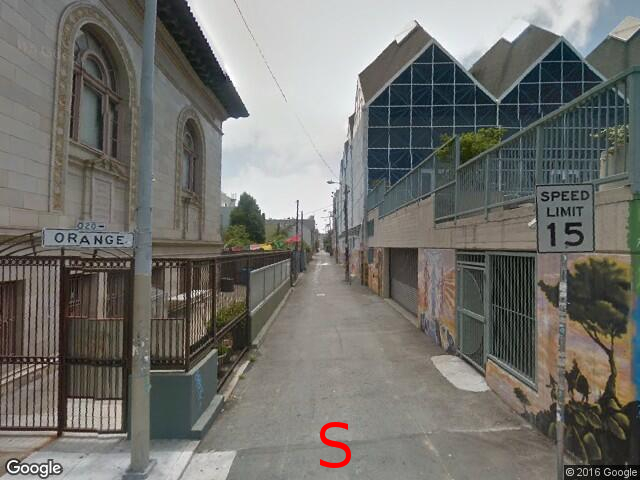}
  \end{subfigure}
  \begin{subfigure}[b]{0.23\textwidth}
    \includegraphics[width=\textwidth]{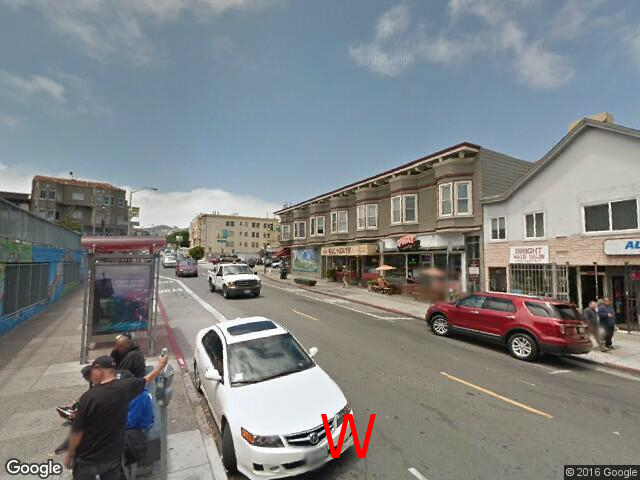}
  \end{subfigure}
  \begin{subfigure}[b]{0.23\textwidth}
    \includegraphics[width=\textwidth]{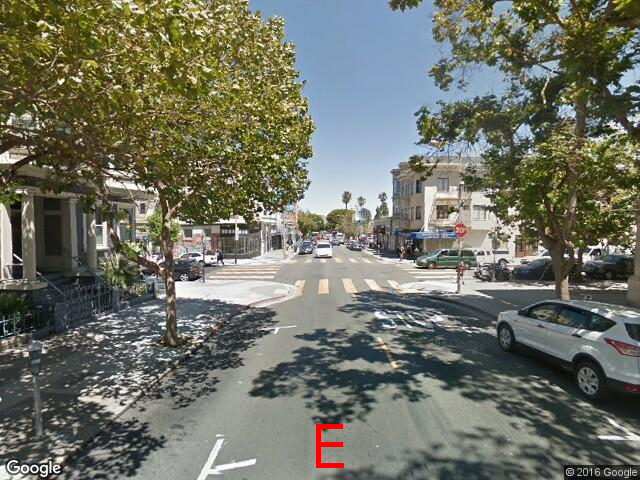}
  \end{subfigure}
\caption{These street-view images are taken from roughly the same location. Which direction do you think the nearest gas station is in?\protect\footnotemark}
\label{fig:1}
\end{figure}
\footnotetext{The correct answer is: W.}
We finely discretize the city into a grid of locations connected by roads. At each location, the agent has access only to street view images pointing towards the navigable directions and it has to pick the next direction to take a step in (Figure~\ref{fig:1}). We show that learning structural characteristics of cities can help the agent reach a destination faster than random walk.\par
Two approaches have been used in the past to achieve this: 1) Reinforcement Learning (RL): a positive reward is associated with each destination location and a negative reward with all other locations. The agent then learns a policy (mapping from state to action) that maximizes the reward expected by executing that policy. In deep RL the policy is encoded by a CNN that outputs the value of performing an action in the current state. A state transition and the observed reward forms a training data-point. Typically, a mini-batch for CNN training is formed by some transitions that are driven by the current policy and some that are random. Some recent works have used this approach to navigate mazes~\cite{async_drl} and small game environments~\cite{drl2}. 2) Supervised Learning: this form of learning requires a large training set of images with labels (e.g. optimal action or distance to nearest destination) that would lead the agent to choose the correct navigable direction at the locations of the images.\par 
Advances in deep CNNs have made it possible to learn high-quality features from images that can be used for various tasks. Most research is focused on recognizing the content of the images e.g. object detection~\cite{imagenet, fast_rcnn, faster_rcnn}, semantic segmentation~\cite{fcn, deeplab, cnn_crf}, edge detection~\cite{deep_edges}, salient area segmentation~\cite{deep_saliency}, etc. However, in robotics and AI one is often required to map image(s) to the choice of an action that a robotic agent must perform to complete a task. For example, the navigation task discussed in this paper requires the CNN to predict the direction of the next step taken by the agent. Such tasks require the CNN to assess the \textit{future implications} of the content of an image, and are relatively unexplored in the literature.\par
We choose supervised learning for this task because of the sparsity of rewards. Out of roughly 100,000 locations in a city grid, only around 30 are destinations (sources of positive reward). RL needs thousands of iterations to sample a transition that ends at a destination (the only time a positive reward happens), especially early in the training process when RL is mostly sampling random transitions. Indeed, Mnih et al.~\cite{async_drl} use 200M training frames to learn navigation in a small synthetic labyrinth. It is not clear if this approach will scale to large-scale environments with highly sparse rewards such as city-scale navigation. On the other hand, there exists an oracle for the problem of navigation in a grid: A* search. A* search finds the shortest path from a starting location to the destination, which gives the optimal action that the agent must perform at every location along the path. Hence it can be used for efficient labelling of large amounts of training images.\par
To summarize, our contributions are:
\begin{itemize}[noitemsep,nolistsep]
\item We collect a dataset of roughly 1M street view images spread across 10 large cities in the USA. This dataset is marked automatically with locations of five types of destinations (Bank of America, church, gas station, high school and McDonald's) using publicly available mapping APIs~\cite{gmaps_nearby, gmaps_roads}
\item We develop and evaluate 3 different CNN architectures that allow an agent to pick a direction at each location to reach the nearest destination. We also compare the performance of these 3 CNN-based models with the model described in~\cite{mcdonalds}, which used hand-crafted features and support vector regression for the same task.
\item We develop a mechanism that uses A* search to generate appropriate labels for our architectures for all the images in the dataset.
\end{itemize}
The rest of the paper is organized as follows: Section~\ref{sec:2} describes the related work in this area, Section~\ref{sec:3} describes our dataset collection process, Section~\ref{sec:4} describes the CNN architectures and training processes and Section~\ref{sec:5} presents results from our algorithm. We discuss the results and conclude in Section~\ref{sec:6}.

\section{Related Work} \label{sec:2}
The computer vision community has explored scene understanding from the perspective of scene classification~\cite{imagenet, deep_scene}, attribute prediction~\cite{deep_scene, attr1, attr2}, geometry prediction~\cite{hoiem_geometry} and pixel level semantic segmentation~\cite{fcn, deeplab, cnn_crf}. All these approaches, however, only reason about information directly present in the scene. Navigating to the nearest destination requires not only understanding the local scene, but also predicting quantities \textit{beyond} the visible scene e.g. distance to nearest destination establishment~\cite{mcdonalds}. Khosla et al.~\cite{mcdonalds} is closest to our paper and addresses the task of navigating to the nearest McDonald's establishment using street-view images. They use a dictionary of spatially pooled Histogram of Oriented Gradient features~\cite{hog} to learn a Support Vector Regressor~\cite{svr} that predicts the distance to the nearest destination in the direction pointed to by the image. In this paper, we first use a CNN to predict the distance and show that data-driven convolutional features perform better than expensive hand crafted features for this task. Next, we propose 2 novel mechanisms to supervise a CNN for this task and show that they lead to better performance. A related line of work deals with image geo-localization~\cite{hays_im2gps, planet} - the problem of localizing the input image in a map. Kendall et al.~\cite{posenet} use CNNs to directly map an input image to the 6D pose of the camera that took the image, in a city-level environment. However, these algorithms only partially solve the problem addressed in this paper - the next steps involve determining the location of the destination and planning a path to it using a map.\par
In artificial intelligence, the problem of picking the optimal action by observing the local surroundings has recently been studied as an application of Deep Reinforcement Learning~\cite{atari}. Works such as~\cite{async_drl, drl2, ai2_deepnav} use Deep RL to learn navigation in artificially generated labyrinths and Minecraft environments. However, the environments are much smaller than the city-level environments considered in this paper and have either repeating artificial patterns~\cite{async_drl} or monotonous non-realistic video game renderings~\cite{drl2}. Another issue with Deep RL is the amount of training data and training epochs required. Even in small-scale environments with denser reward-yielding locations compared to our environments, Mnih et al.~\cite{async_drl} use 50 training epochs with each epoch made of 4M training frames, while Oh et al.~\cite{drl2} use up to 200 epochs. In contrast, our algorithms require 8 epochs to train, while our initialization network~\cite{vgg} requires 74 training epochs through 1M images. To our knowledge, no deep RL algorithm has attempted the task of learning to navigate in city-scale environments with real-world noisy imagery. Mirowski et al.~\cite{deepmind_deepnav} use auxiliary tasks like depth prediction from RGB and loop closure detection to alleviate the sparse rewards problem. However, their experiments are performed in artificially generated environments much smaller than the city-scale environments we operate in.\par
In robotics, active vision~\cite{active_vision} has been used to control a robot to reach a destination based on local observations. The use of shortest paths to train classifiers that map the state of an agent (encoded by local observations) to action first appears in~\cite{burgard_active}. They train a decision tree on various hand-crafted attributes of locations in a supermarket (e.g. type of aisle, visible products, etc.) to control a robot to reach a target product efficiently. Aydemir et al.~\cite{aydemir_active} use a chain graph model that relies on object-object co-occurrence and object-room type co-occurrence to control a robot to reach a destination in a 3D indoor environment.
\section{Dataset} \label{sec:3}
\begin{figure}
\includegraphics[width=0.49\textwidth]{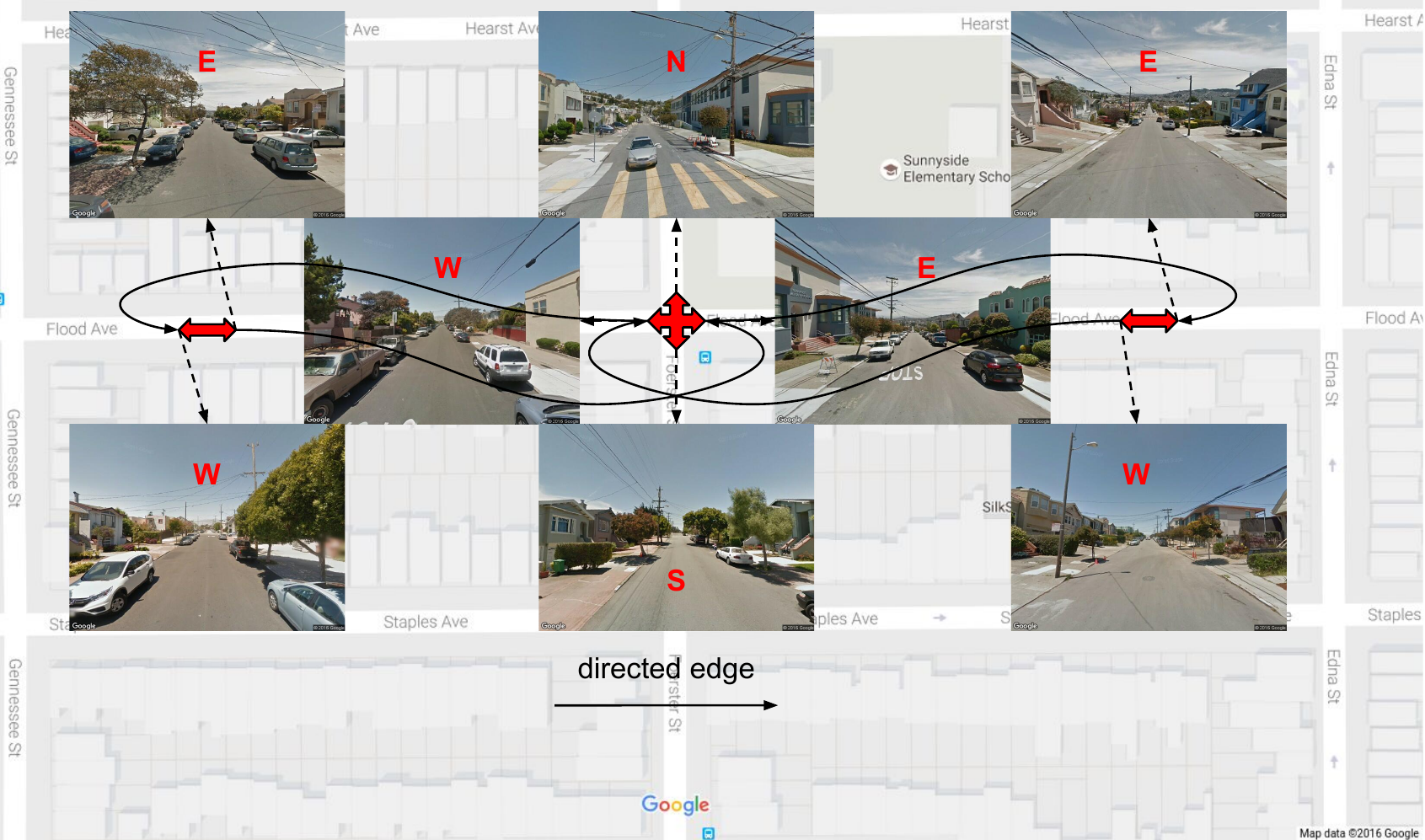}
\caption{Directed graph illustration. The red arrows represent nodes, encoded by location and direction. Each node has an associated street-view image, taken at the node's location and pointing in the node's direction. Nodes are connected by roads (solid connectors in the figure).}
\label{fig:2}
\end{figure}
\begin{figure}
\centering
  \begin{subfigure}[b]{0.23\textwidth}
    \includegraphics[width=\textwidth]{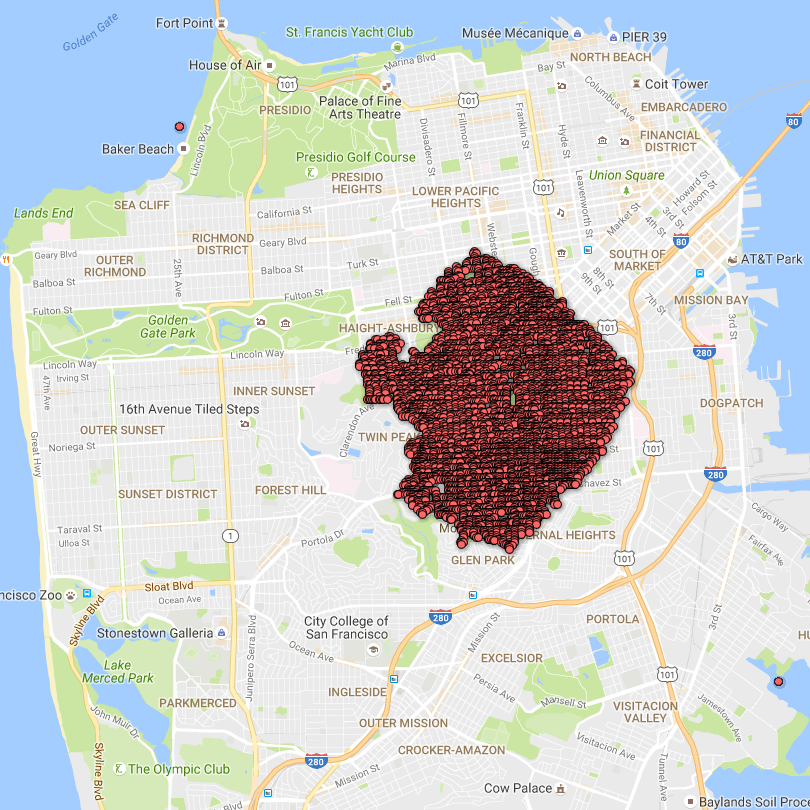}
  \end{subfigure}
  \begin{subfigure}[b]{0.23\textwidth}
    \includegraphics[width=\textwidth]{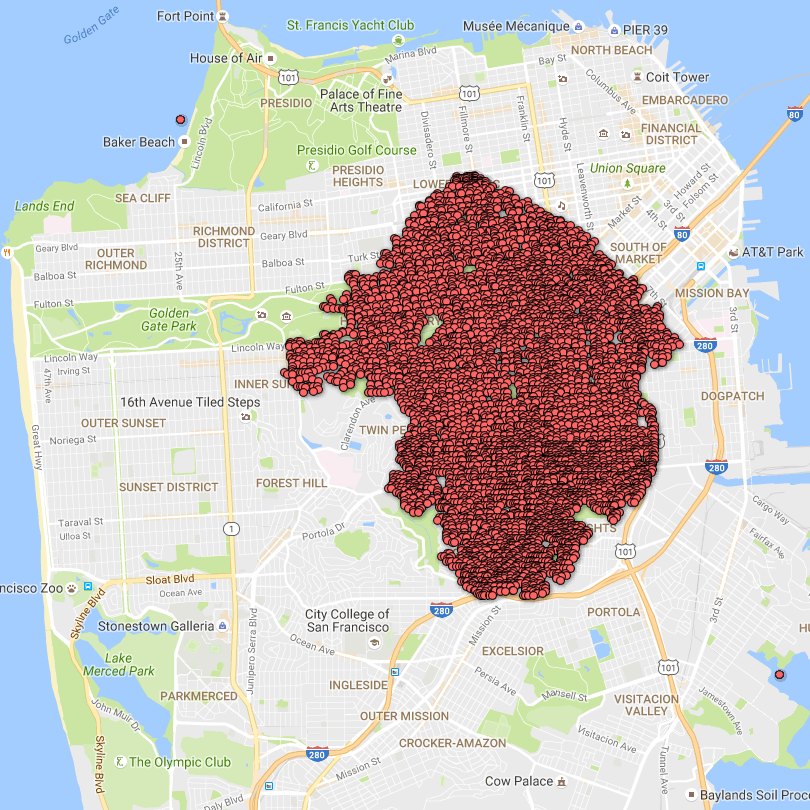}
  \end{subfigure}
  \begin{subfigure}[b]{0.23\textwidth}
    \includegraphics[width=\textwidth]{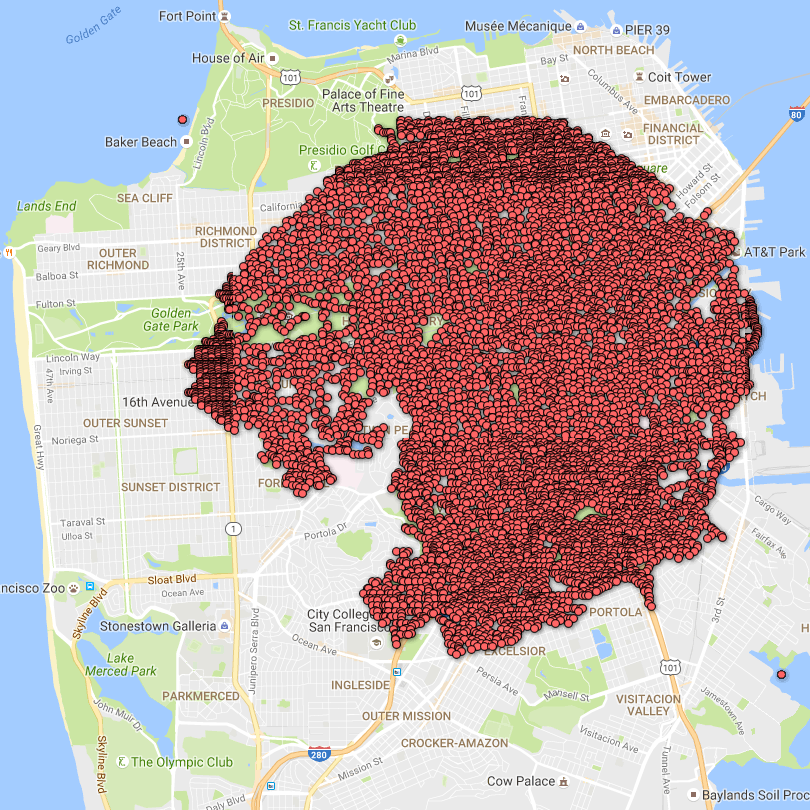}
  \end{subfigure}
  \begin{subfigure}[b]{0.23\textwidth}
    \includegraphics[width=\textwidth]{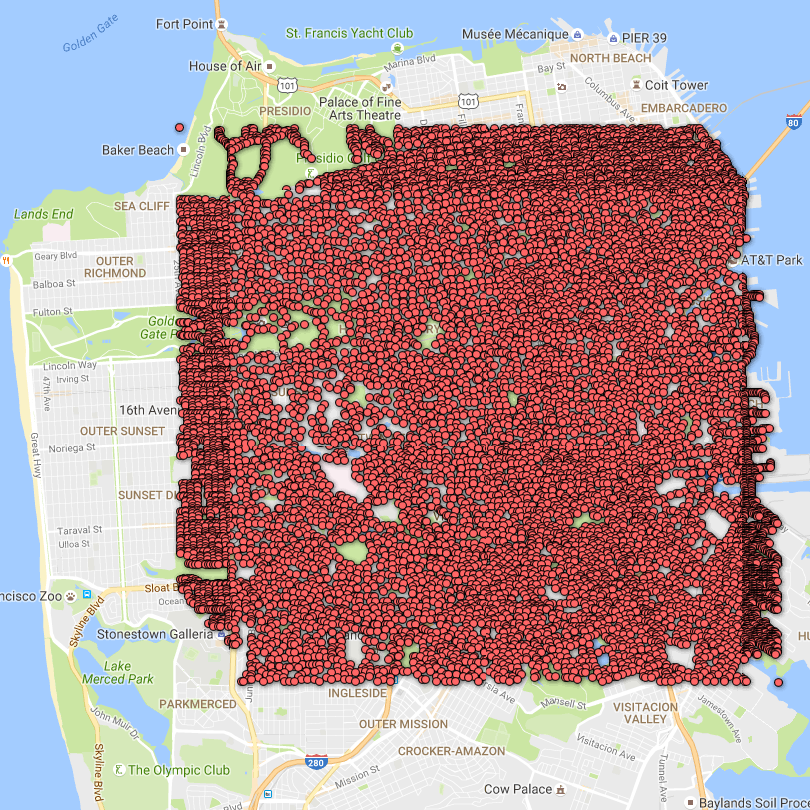}
  \end{subfigure}
\caption{Progression of breadth-first enumeration for constructing the San Francisco city graph.}
\label{fig:3}
\end{figure}
The DeepNav agent has a location and a heading associated with it, and traverses a directed graph covering the city. Figure~\ref{fig:2} shows a visualization of a small part of this graph. Nodes are defined by the tuple of street view image location (latitude, longitude) and direction (North, South, East, and West). Hence each location can host upto 4 nodes. Edges in this graph represent a one-way road i.e. a node is connected to a neighbouring node by a directed edge if there is a road that allows the agent to travel from the first node to the second node. However, edges only connect neighbouring nodes facing the same direction. Hence travelling along an edge allows the agent to take one step in the direction of its current heading. To allow the agent to turn in place, all nodes at one location are cyclically connected with bidirectional edges. Lastly, a node exists only if it is connected to a node at a different location. This implies that locations along a road only have 2 nodes per location, while intersections have 3 or 4 nodes per location depending on the type of intersection. At intersections of more than 4 roads, we ignore all roads that do not point in the cardinal directions. This construction makes it possible for the agent to travel between any two nodes in the graph. Each node has a 640 x 480 image cropped from the Google Street View panorama at the location of the node. This image has a field-of-view of $90\degree$ and points in the direction associated with the node. While cardinal directions N, S, E, W are shorthand, the street-view crops have continuous direction consistent with the road direction.\par
To control the granularity of node locations, we tessellate the city limits into square bins of side 25m and consider the centers of these bins to be node locations. All street-view panorama locations that fall inside a bin are snapped to the center of the bin. However, the actual street view images are captured from the edges of the bins, to ensure visual continuity.\par
We generate one such graph per city. First the limits are specified by the latitude and longitude of two opposite corners of a rectangular region. We then start a breadth-first enumeration of the locations in the city, starting at the center of the rectangle (shown in Figure~\ref{fig:3} for San Francisco.). This enumeration stops when the specified geographical limits are reached. Table~\ref{tab:1} shows the number of images in the graphs of the 10 cities in our dataset.
\begin{figure}
\centering
\includegraphics[width=0.49\textwidth]{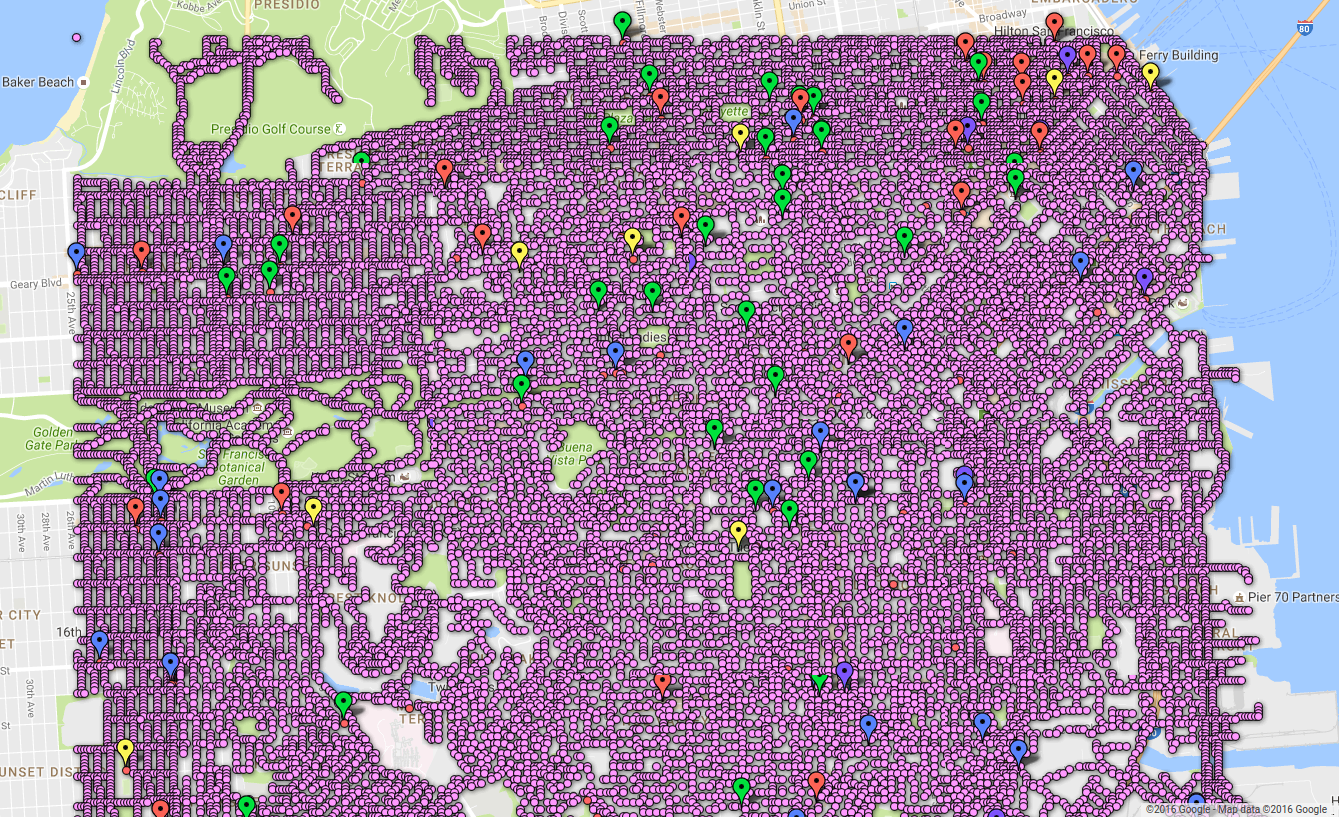}
\caption{San Francisco graph locations and destinations (red: Bank of America, green: church, blue: gas station, yellow: high school, purple: McDonald's)}
\label{fig:4}
\end{figure}
\begin{table}
\centering
\resizebox{0.45\textwidth}{!}{%
\begin{tabular}{|c|c|ccccc|}
\hline
\textbf{City} & \textbf{Images} & \textbf{BofA} & \textbf{church} & \textbf{gas station} & \textbf{high school} & \textbf{McDonald's}\\
\hline
Atlanta & 78,808 & 10 & 32 & 32 & 7 & 7\\
Boston & 105,000 & 40 & 40 & 39 & 20 & 20\\
Chicago &	105,001 & 22 & 33 & 10 & 15 & 32\\
Dallas & 105,000 & 7 & 25 & 35 & 9 & 13\\
Houston &	117,297 & 8 & 19 & 30 & 4 & 14\\
Los Angeles &	80,701 & 9 & 15 & 30 & 6 & 13\\
New York & 105,148 & 30 & 20 & 21 & 27 & 31\\
Philadelphia & 105,000 & 14 & 42 & 35 & 30 & 19\\
Phoenix &	101,419 & 4 & 23 & 29 & 18 & 15\\
San Francisco &	101,741 & 35 & 50 & 45 & 22 & 12\\
\hline
\textbf{Total} & \textbf{1,005,115} & \textbf{179} & \textbf{299} & \textbf{306} & \textbf{158} & \textbf{176}\\
\hline
\end{tabular}}
\caption{Number of images and destinations in city graphs}
\label{tab:1}
\end{table}
\subsection{Destinations}
We consider 5 classes of destinations: Bank of America, church, gas station, high school and McDonald's. These were chosen because of their ubiquity and distinguished visual appearance. Given the city graph limits we use Google Maps nearby search~\cite{gmaps_nearby} with an appropriate radius to find the locations of all establishments of these classes in the city. Next, we use the Google Maps Roads API~\cite{gmaps_roads} to snap these locations to the nearest road location. This is necessary because these establishments are often large in size and street-view images exist only along roads. Figure~\ref{fig:4} shows the destination locations for San Francisco, while Table~\ref{tab:1} shows the number of destinations in each city graph found by our program.
\section{CNN Architecture and Training} \label{sec:4}
Given the city graph and street-view images, we want to train a convolutional neural network to learn the visual features that are common across paths leading to various destinations. We propose 3 methods to label the training images (and corresponding inference algorithms) to accomplish this. The first approach, DeepNav-distance, trains the network to estimate the distance to the nearest destination in the direction pointed to by the training image. The second approach, DeepNav-direction, learns a mapping between a training image and the optimal action to be performed at the image location. The third approach, DeepNav-pair, decomposes the problem of picking the optimal action into pairs of decisions, and employs a Siamese CNN architecture.
\begin{figure*}
	\centering
	\begin{subfigure}[b]{0.49\textwidth}
		\includegraphics[width=0.67\textwidth]{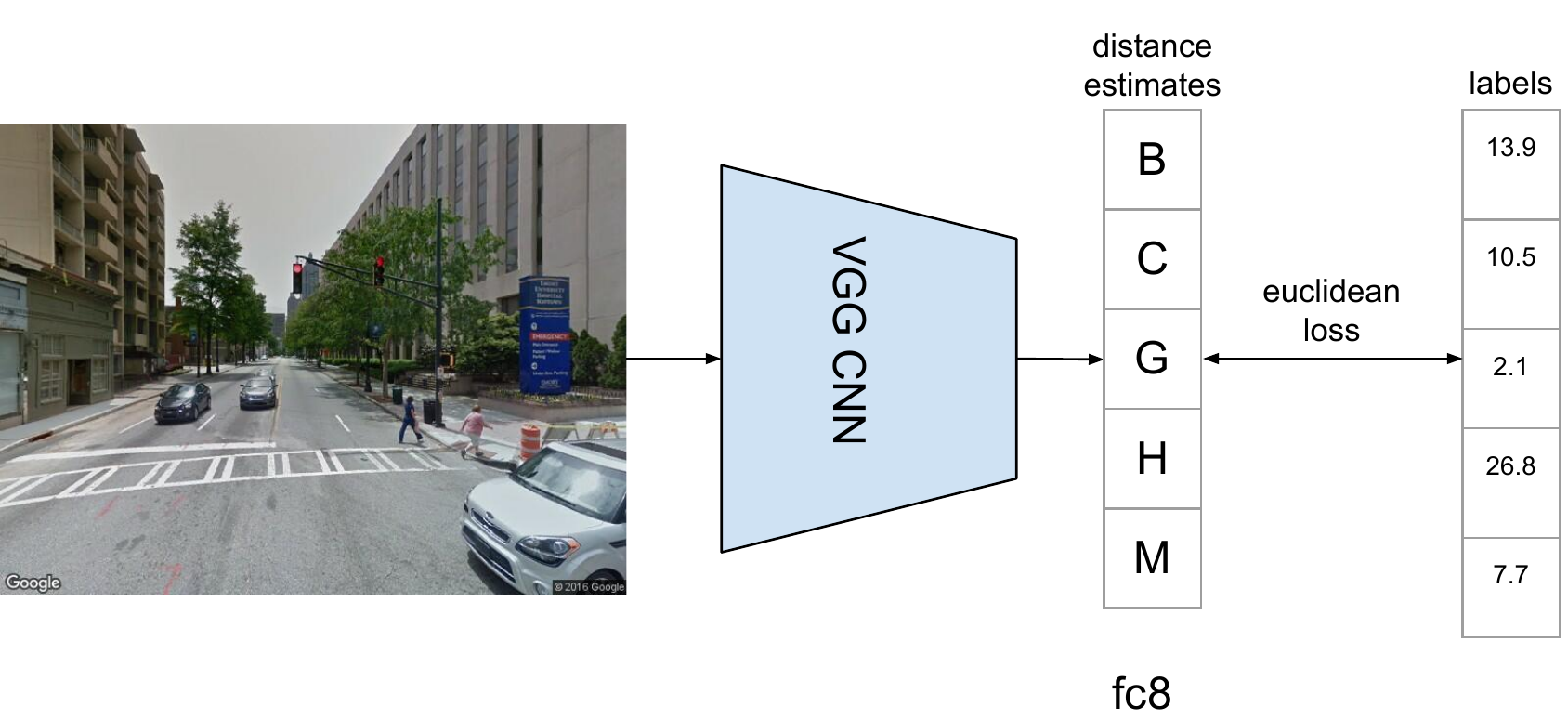}
		\includegraphics[width=\textwidth]{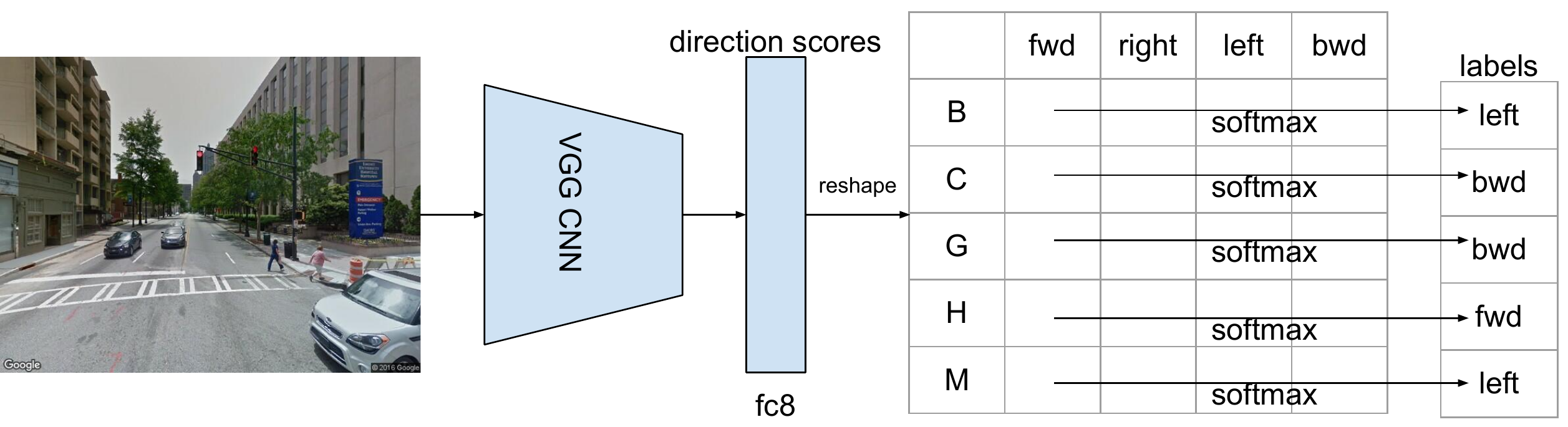}
		\caption{DeepNav-distance (top), DeepNav-direction (bottom)}
		\label{fig:3a}
	\end{subfigure}
	\begin{subfigure}[b]{0.49\textwidth}
		\includegraphics[width=\textwidth]{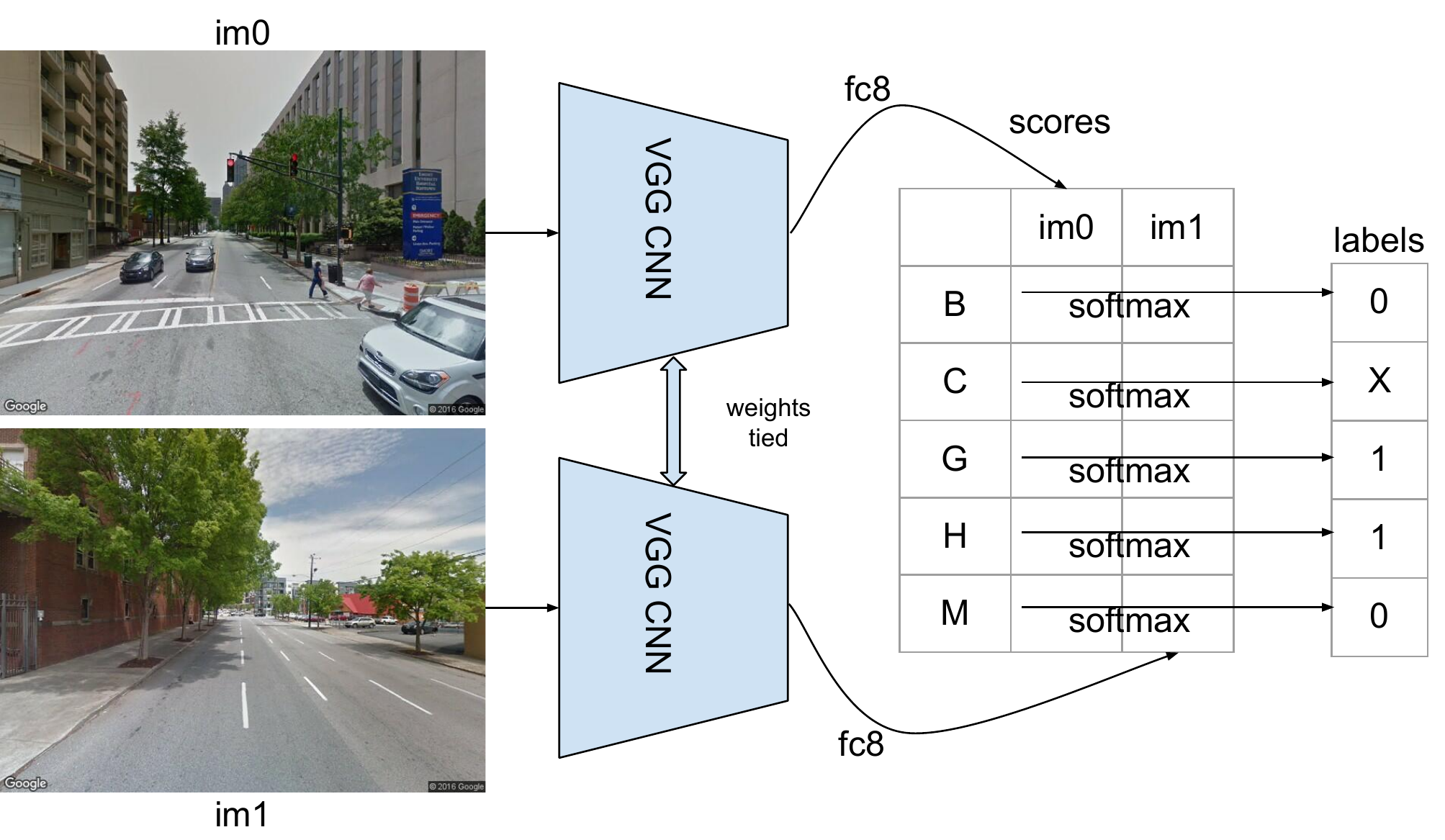}
		\caption{DeepNav-pair}
		\label{fig:3b}
	\end{subfigure}
	\caption{DeepNav CNN architectures. Abbreviations for destinations: B = Bank of America, C = Church, G = Gas station, H = High school, M = McDonald's.}
\end{figure*}
\subsection{DeepNav-distance}
In this scheme we label each image with the square-root of the straight-line distance from the image location to the nearest destination establishment, in the $90\degree$ arc corresponding to the direction of the image. We collect 5 labels corresponding to 5 destination classes for each image. To train DeepNav-distance, we modify the last fully connected layer (fc8) of the VGG 16-layer network~\cite{vgg} to have 5 output units (see Figure~\ref{fig:3a}). The objective function minimized by this algorithm is the Euclidean distance between the fc8 output and the 5-element label vector. If a particular node has no destination establishment of a category in its arc, the corresponding element of the label vector is set to a high value that is ignored by the objective function. Hence, a training image is used for learning as long as it has at least one kind of destination in its arc.\par
We use a greedy approach at test time: we forward images from all available directions at the current location of the agent through the CNN. The agent takes a step in the direction that is predicted by the CNN to have the least distance estimate. This approach is inspired from~\cite{mcdonalds}, and is intended to investigate the change in performance by using an end-to-end convolutional neural network pipeline instead of hand-crafted features and support vector regression.
\subsection{DeepNav-direction}
This approach learns to map an input image to the optimal action to be performed at that particular location and direction. The graph allows the agent to perform up to 4 actions at a node: move forward, move backward, move left or move right (the last 3 are composed of the primitive actions of moving forward and turning in place). We note that A* search in the graph finds the shortest path from any starting location to a destination, and hence can generate optimal action labels for each node location along the shortest path. For example, if the A* path at a node turns East, the image at that node facing East is labelled `move forward', the image facing North is labelled `move right', and so on. Algorithm~\ref{alg:1} describes the process of generating labels for all training images using A* search, for one class of destination (e.g. high schools). The algorithm is repeated for all 5 classes of destinations to get 5 optimal action labels for each training image. Each label can take one of four values.
\begin{algorithm}
	\KwData{City graph $G$, destinations $D$}
	\KwResult{Optimal action labels for each node}
	\While{$\exists$ unlabelled node}{
		$n \gets$ unlabelled node\;
		$\mathrm{shortest\_path} \gets []$\;
		$\mathrm{min\_cost} \gets \infty$\; 
		\ForEach{$d \in D$}{
			$cost, path \gets \mathrm{A*}(n, d, G)$\;
			\If{$cost < \mathrm{min\_cost}$}{
				$\mathrm{min\_cost} \gets cost$\;
				$\mathrm{shortest\_path} \gets path$\;
			}
		}
		\ForEach{node $i \in \mathrm{shortest\_path}$}{
			label each node at $i.location$ using A* action\;
		}
 	}
 	\caption{Generating labels for DeepNav-direction}
 	\label{alg:1}
\end{algorithm}
To train DeepNav-direction, we modify the last fully connected layer (fc8) of the VGG 16-layer network~\cite{vgg} to have 20 output units (see Figure~\ref{fig:3a}). These 20 outputs are interpreted as scores for the 4 possible actions (along the columns) for the 5 destination classes (along the rows). The objective function minimized by DeepNav-direction is the softmax loss computed independently for each destination class. At test time, the image from the current position and direction of the agent is forwarded through the convolutional neural network, and the agent performs the highest scoring available action.
\subsection{DeepNav-pair}
This approach also learns to select the optimal action to be performed at a particular location and direction like DeepNav-direction, but through a different formulation. DeepNav-direction gets only the forward-facing image as input, and does not get to `see' in all directions before choosing an action. This is an important action performed by a variety of animals including primates, birds and fish while navigating unknown environments~\cite{psych}. This behaviour can be implemented by the Siamese architecture shown in Figure~\ref{fig:3b}. We enumerate all pairs of images at a location, and use the optimal action given by A* to label at most one image from each pair as the `favorable' image. A pair is ignored if it does not contain a favorable image. For example, if the A* path at a node turns East, the second image in the North-East pair is marked favorable, while the North-South pair is ignored. Algorithm~\ref{alg:2} shows the process of gathering the labels for all such pairs in the training dataset, and it is repeated for each destination class.
\begin{algorithm}
	\KwData{City graph $G$, destinations $D$}
	\KwResult{Optimal action labels for each image-pair, where pairs are formed between images at a common location}
	\While{$\exists$ unlabelled node}{
		$n \gets$ unlabelled node\;
		$\mathrm{shortest\_path} \gets []$\;
		$\mathrm{min\_cost} \gets \infty$\; 
		\ForEach{$d \in D$}{
			$cost, path \gets \mathrm{A*}(n, d, G)$\;
			\If{$cost < \mathrm{min\_cost}$}{
				$\mathrm{min\_cost} \gets cost$\;
				$\mathrm{shortest\_path} \gets path$\;
			}
		}
		\ForEach{node $i \in \mathrm{shortest\_path}$}{
			\ForEach{pair $p \in pairs(i.location)$}{				
				\uIf{$direction(p.first) ==$ A* direction}{
					$(p.first, p.second) \gets$ label 0\;
				}\uElseIf{$direction(p.second) ==$ A* direction}{
					$(p.first, p.second) \gets$ label 1\;
				}\Else(ignore pair){
					$(p.first, p.second) \gets$ label X\;
				}
			}
		}
 	}
 	\caption{Generating labels for DeepNav-pair}
 	\label{alg:2}
\end{algorithm}
To train DeepNav-pair, we create a Siamese network with 2 copies of the DeepNav-distance network, as shown in Figure~\ref{fig:3b}. The outputs of the fc8 layer are treated as scores instead of distance estimates, and stacked as columns. A softmax loss is applied across the columns, independently for each destination. Hence the network learns to pick the image pointing in the direction of the optimal action, from all existing images at a location. At test time, we keep only one branch of the Siamese architecture and use the fc8 outputs as a score. All images from the current location of the agent are forwarded through the network, and the agent takes a step in the direction of the image that has the highest score predicted by the network.
\subsection{Training}
We train the convolutional neural networks using stochastic gradient descent (SGD) implemented in the Caffe~\cite{caffe} library. The learning rate for SGD starts at $10^{-3}$ for DeepNav-pair and DeepNav-direction, and $10^{-4}$ for DeepNav-distance. All models are trained for 8 epochs, with the learning rate dropping by a factor of 10 after the 4th and 6th epochs. We set the weight decay parameter to $5*10^{-4}$ and SGD momentum to 0.9. Training on an NVIDIA TITAN X GPU takes roughly 72 hours for each DeepNav model. All networks are initialized from the public VGG 16-layer network~\cite{vgg} except the fc8 layers, which are initialized using the Xavier~\cite{xavier_init} method.
\subsubsection{Geographically weighted loss function}
The DeepNav models are being trained to identify visual features that indicate the path to a destination. We expect these visual features to be concentrated around destination locations. To relax the CNN loss function for making a wrong decision based on low-information visual features far away from the destinations, we modify the loss function by weighing the training samples geographically. Specifically, the weight for the training sample reduces as length of the shortest path from its location to a destination increases. The geographically weighted loss function $L_g$ for a mini-batch of size $N$ is constructed from the original loss function $L$:
\begin{equation}
L_g = \sum_{i=1}^{N}\lambda^{l_i}L_i
\end{equation}
where $0 < \lambda < 1$ is the geographic weighting factor. We apply geographic weighting to the loss functions for DeepNav-direction and DeepNav-pair with $\lambda = 0.9$. In our experiments, we observe that SGD training for these networks does not converge without geographic weighting. We do not apply geographic weighting to DeepNav-distance because it is penalized less for predicting a slightly wrong distance estimate far away from the destination by use of square-root of the distance as the label.
\section{Results} \label{sec:5}
In this section, we evaluate the ability of the various DeepNav models to navigate unknown cities and reach the nearest destination and compare them with the algorithm presented in~\cite{mcdonalds}. For reference, we also present the metrics for A* search - note that A* search has access to the entire city graph and destination location while planing the path, while the other methods only have access to images from the agent's current location.
\subsection{Baselines}
The algorithm for navigating to the nearest McDonald's presented by Khosla et al. in~\cite{mcdonalds} serves as our first baseline. This algorithm extracts Histogram of Oriented Gradient~\cite{hog} features densely over the entire image and applies K-means to learn a dictionary of size 256. It then uses locality-constrained linear coding~\cite{llc} to assign the descriptors to the dictionary in a soft manner, and finally builds a 2-level spatial pyramid~\cite{spp} to obtain a final feature of dimension 5376. We use the publicly available code from the authors~\cite{fe_code} to compute the features. To speed up dictionary creation, we create it from a collection of 18,000 images sampled randomly from all the training cities (3000 from each city). A Support Vector Regressor (SVR)~\cite{svr, liblinear} is learnt to map an input image to the square root of the distance to the nearest destination establishment in the $90\degree$ arc in the direction pointed to by the node. We chose the regularization constant in the SVR by picking the value which minimized the Euclidean error over the 4 test cities (see next section). Our second baseline is a random walk algorithm. This algorithm picks a random action at each node.
\subsection{Experimental setup} \label{sec:5-2}
We train the DeepNav models on 6 cities (Atlanta, Boston, Chicago, Houston, Los Angeles, Philadelphia) and test them on the held-out 4 cities (Dallas, New York, Phoenix, San Francisco). This avoids the bias of training and testing on disjoint parts of the same city, and tests whether the algorithms are able to learn about structure in man-made environments (cities) and use that knowledge in unknown environments. For each test city, we sample 10 starting locations uniformly around each destination with an average path length $d_s$. Thus we get around 250 starting locations for each city. All our agents start at these locations facing a random direction and navigate the city using the inference procedures described in Section~\ref{sec:4}. To prevent looping, the agent is not allowed pick the same action twice from a node. If the agent has no option to move from a location, it is re-spawned at the nearest node with an open option to move. The agent is considered to have reached a destination if it visits a node within 75m of it, the maximum number of steps is set to 1000.
\begin{table}[t!]
\centering
\resizebox{0.49\textwidth}{!}{
\begin{tabular}{|c|c|c|c|}
\hline
\multirow{2}{*}{\textbf{Method}} & \multicolumn{3}{c|}{\textbf{Expected number of steps}}\\
\cline{2-4}
& $d_s$=470m & $d_s$=690m & $d_s$=970m\\
\hline
Random walk & 733.99 & 854.89 & 911.85\\
A* & 18.73 & 27.32 & 39.57\\
\hline
HOG+SVR~\cite{mcdonalds} & 588.66 & 705.31 & 791.93\\
DeepNav-distance & 580.69 & \textbf{684.22} & 773.02\\
DeepNav-direction & 626.28 & 697.26 & 780.53\\
DeepNav-pair & \textbf{553.39} & 689.04 & \textbf{766.32}\\
\hline
\end{tabular}}
\caption{Expected number of steps for various algorithms.}
\label{tab:2}
\end{table}
\subsection{Evaluation metrics}
We use the two metrics proposed in~\cite{mcdonalds}: 1) success rate (fraction of times the agent reaches its destination) and 2) average number of steps taken to reach the destination in the successful trials.  To ease comparison of various methods, we propose the expected number of steps metric, which is calculated as $s*L + (1-s)*L_{max}$ where $s$ is the success rate, $L$ is the average number of steps for successful trials and $L_{max}$ is the maximum number of steps (1000 in our case).\par
The metrics are averaged over all the starting locations of a city (and over 20 trials for the random walker). Table~\ref{tab:2} shows the expected number of steps averaged over all destinations and starting locations, for $d_s = 470m$, $690m$ and $970m$. DeepNav-pair outperforms the baseline as well as other DeepNav architectures for most starting distances. We hypothesize that DeepNav-pair can perform better because it is the only algorithm that is trained by `looking' in all directions and choosing the optimal direction. We also note that DeepNav-distance outperforms the agent from~\cite{mcdonalds}, indicating the better quality of deep features. Tables~\ref{tab:3} and~\ref{tab:4} show the success rate and the average number of steps for successful trials for $d_s = 480m$. We see that DeepNav-pair has the highest average success rate. DeepNav-direction has the lowest average path length for successful trials, but lowest success rate. This indicates that it is effective only for short distances. Detailed metrics for $d_s = 690m$ and $970m$ are presented in Tables~\ref{tab:5}-\ref{tab:8}.\par
\begin{figure*}[p!]
\centering
  \begin{subfigure}[b]{0.19\textwidth}
    \includegraphics[width=\textwidth]{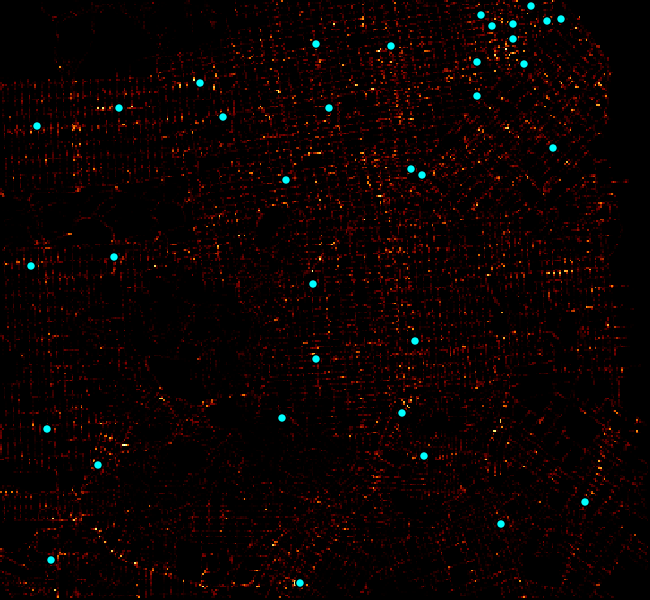}
    \caption{DeepNav-pair}
  \end{subfigure}
  \begin{subfigure}[b]{0.19\textwidth}
    \includegraphics[width=\textwidth]{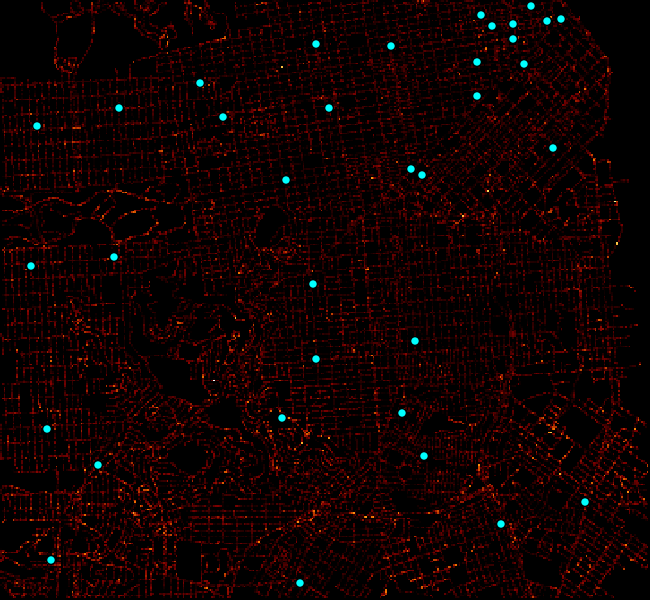}
    \caption{DeepNav-direction}
  \end{subfigure}
  \begin{subfigure}[b]{0.19\textwidth}
    \includegraphics[width=\textwidth]{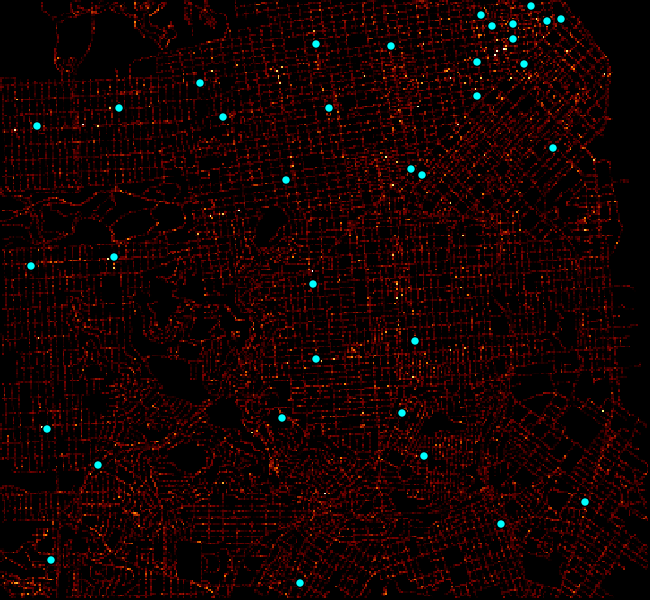}
    \caption{DeepNav-distance}
  \end{subfigure}
  \begin{subfigure}[b]{0.19\textwidth}
    \includegraphics[width=\textwidth]{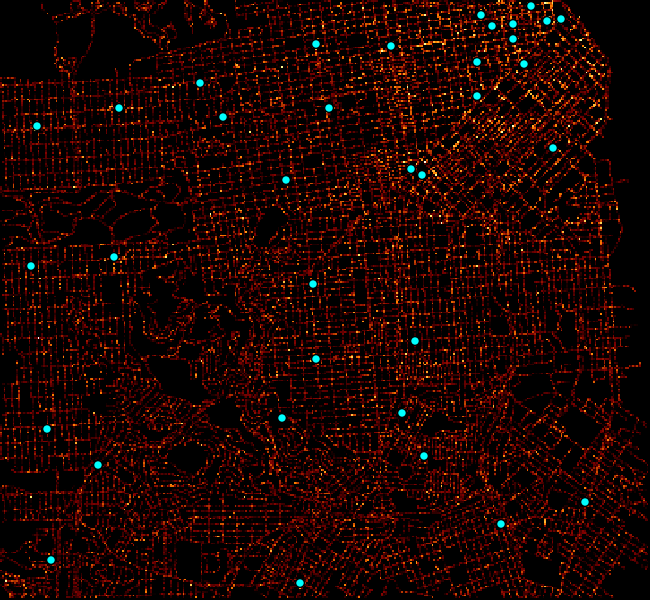}
    \caption{HOG+SVR~\cite{mcdonalds}}
  \end{subfigure}
  \begin{subfigure}[b]{0.19\textwidth}
    \includegraphics[width=\textwidth]{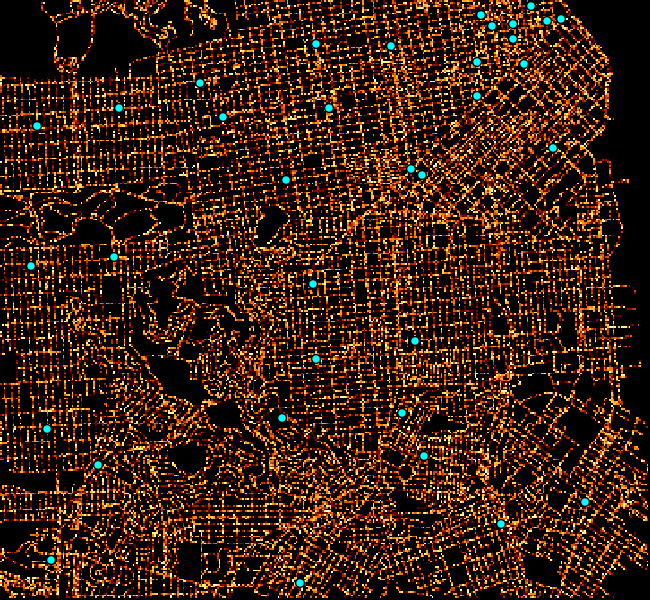}
    \caption{Random walk}
  \end{subfigure}
\caption{Confidence of predictions while navigating for Bank of America (blue dots) in San Francisco (a test city). Brighter colors imply higher variance. Concentration of high-variance regions indicates that DeepNav-pair confidence increases near destinations and it effectively learns visual features common to optimal paths.}
\label{fig:6}
\end{figure*}
\begin{figure*}[p!]
\centering
  \begin{subfigure}[b]{0.19\textwidth}
    \includegraphics[width=\textwidth]{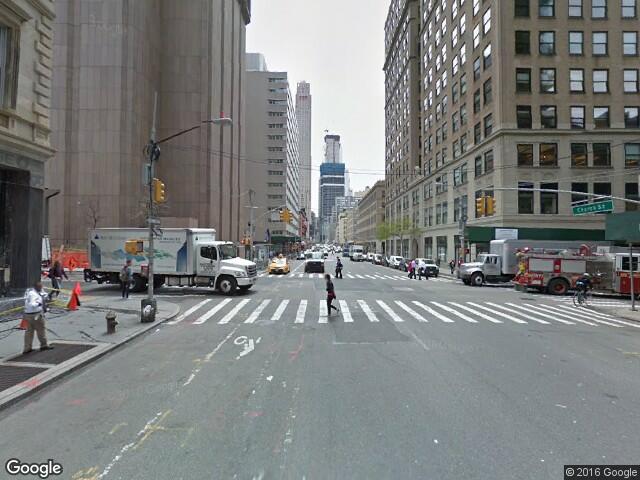}
  \end{subfigure}
  \begin{subfigure}[b]{0.19\textwidth}
    \includegraphics[width=\textwidth]{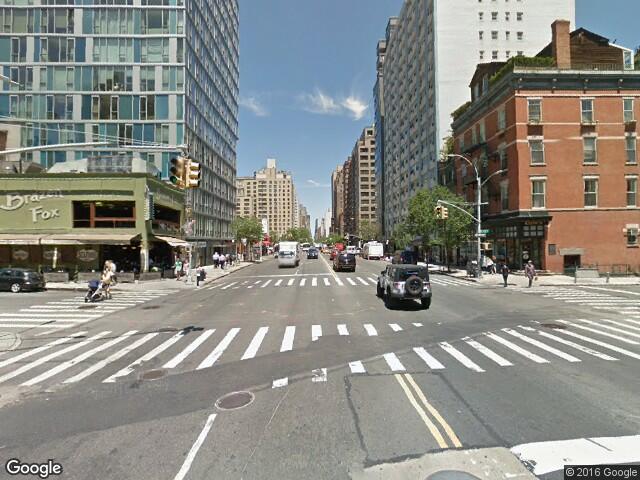}
  \end{subfigure}
  \begin{subfigure}[b]{0.19\textwidth}
    \includegraphics[width=\textwidth]{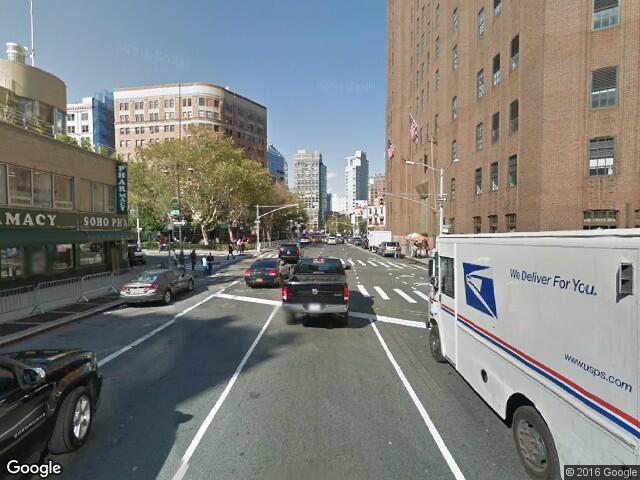}
  \end{subfigure}
  \begin{subfigure}[b]{0.19\textwidth}
    \includegraphics[width=\textwidth]{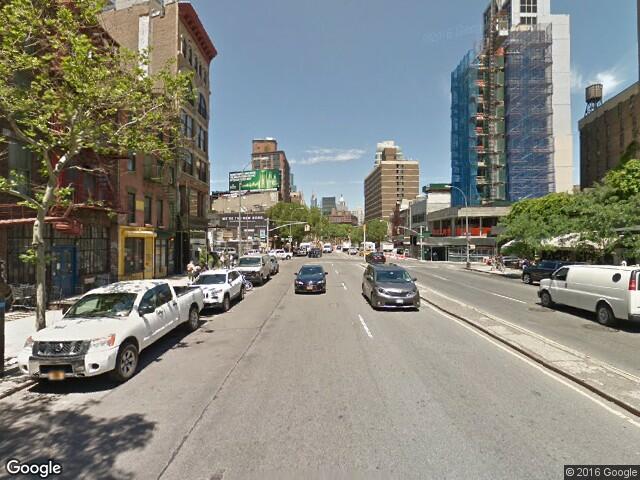}
  \end{subfigure}
  \begin{subfigure}[b]{0.19\textwidth}
    \includegraphics[width=\textwidth]{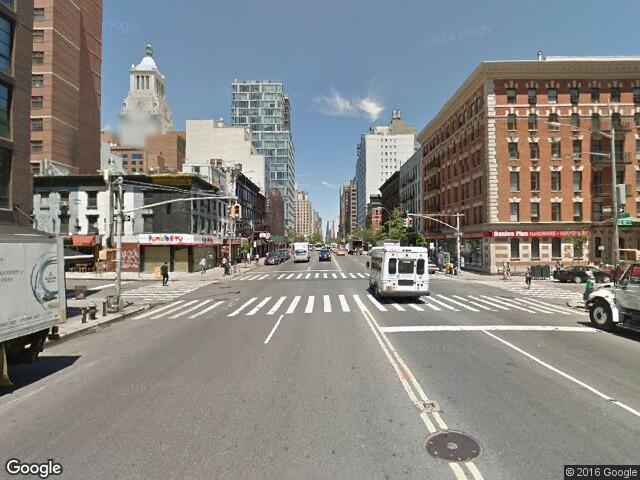}
  \end{subfigure}
  \begin{subfigure}[b]{0.19\textwidth}
    \includegraphics[width=\textwidth]{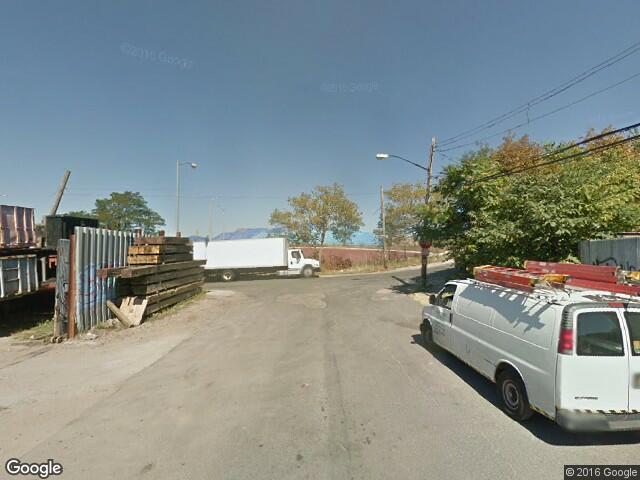}
  \end{subfigure}
  \begin{subfigure}[b]{0.19\textwidth}
    \includegraphics[width=\textwidth]{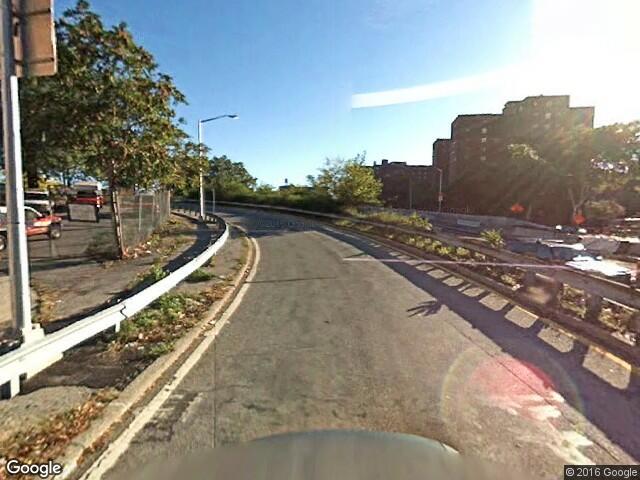}
  \end{subfigure}
  \begin{subfigure}[b]{0.19\textwidth}
    \includegraphics[width=\textwidth]{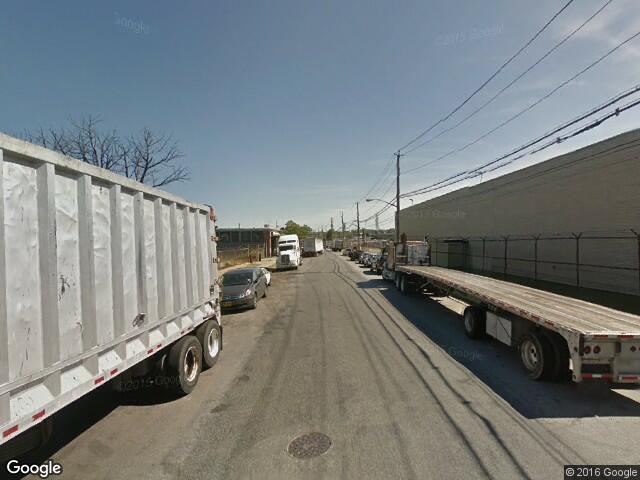}
  \end{subfigure}
  \begin{subfigure}[b]{0.19\textwidth}
    \includegraphics[width=\textwidth]{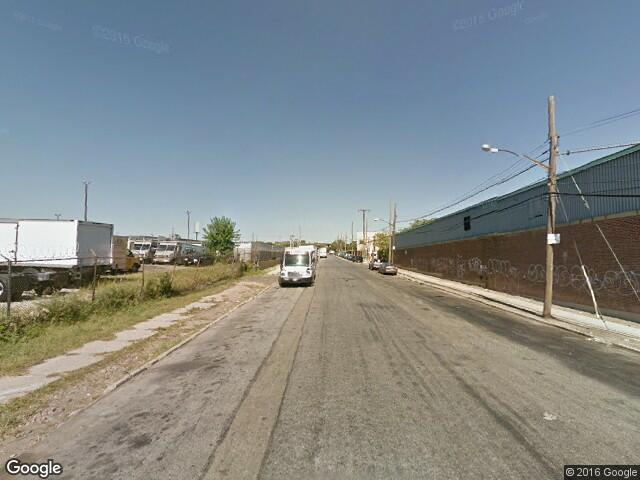}
  \end{subfigure}
  \begin{subfigure}[b]{0.19\textwidth}
    \includegraphics[width=\textwidth]{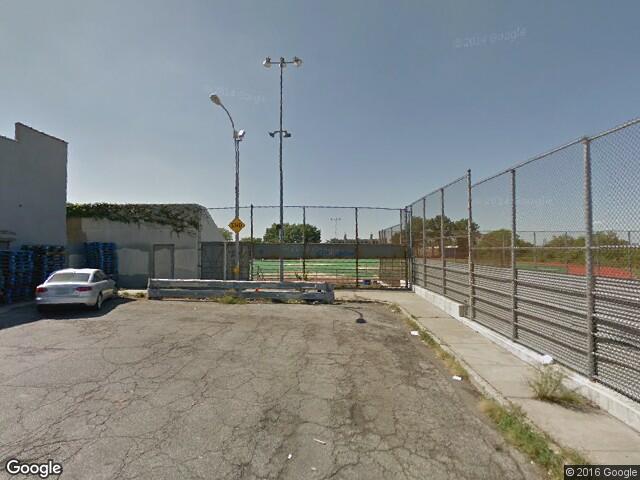}
  \end{subfigure}
  \hrule
  \begin{subfigure}[b]{0.19\textwidth}
    \includegraphics[width=\textwidth]{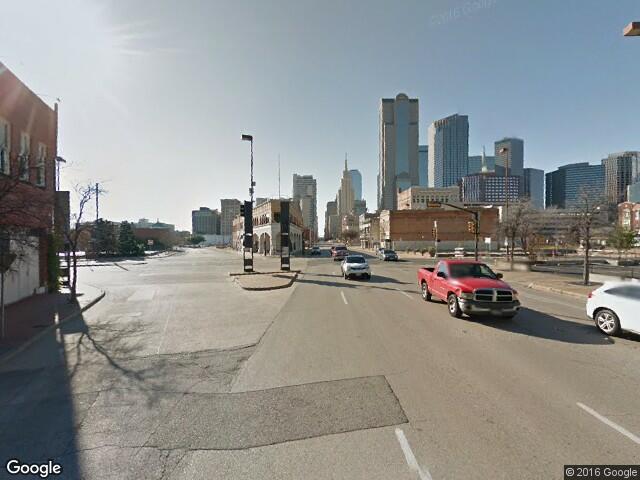}
  \end{subfigure}
  \begin{subfigure}[b]{0.19\textwidth}
    \includegraphics[width=\textwidth]{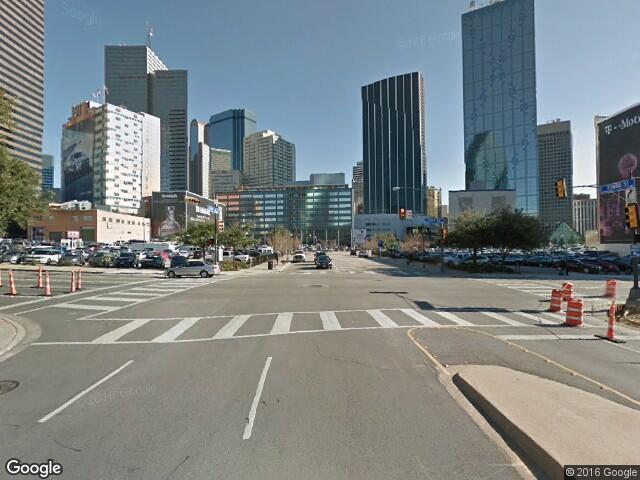}
  \end{subfigure}
  \begin{subfigure}[b]{0.19\textwidth}
    \includegraphics[width=\textwidth]{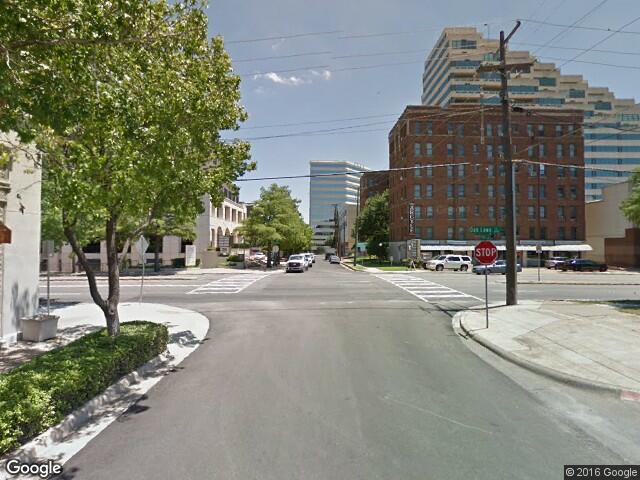}
  \end{subfigure}
  \begin{subfigure}[b]{0.19\textwidth}
    \includegraphics[width=\textwidth]{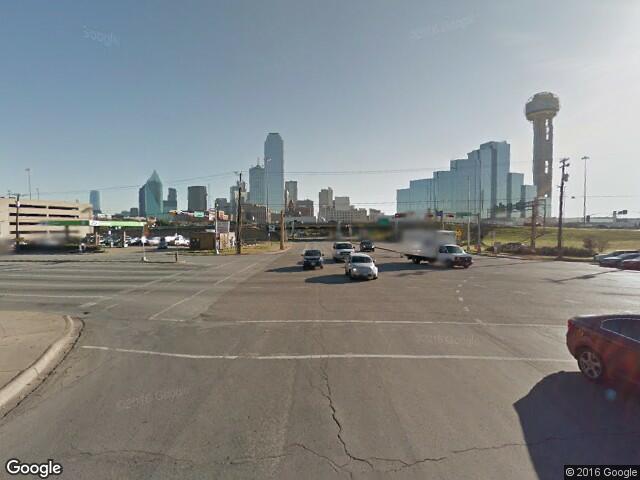}
  \end{subfigure}
  \begin{subfigure}[b]{0.19\textwidth}
    \includegraphics[width=\textwidth]{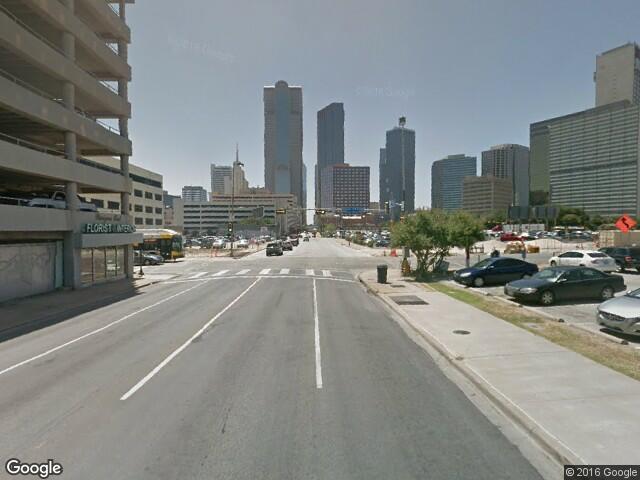}
  \end{subfigure}
  \begin{subfigure}[b]{0.19\textwidth}
    \includegraphics[width=\textwidth]{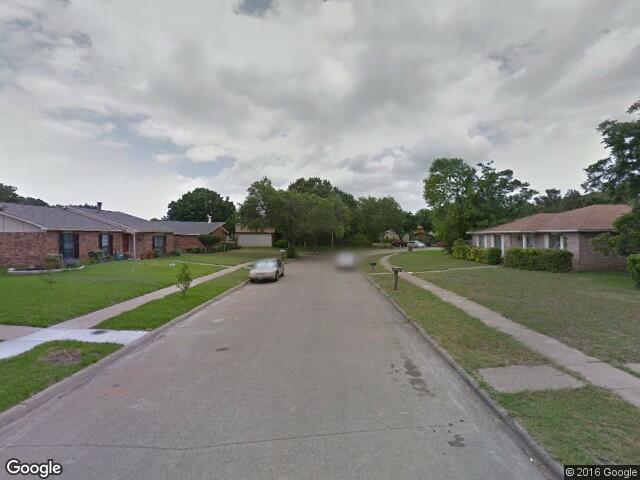}
  \end{subfigure}
  \begin{subfigure}[b]{0.19\textwidth}
    \includegraphics[width=\textwidth]{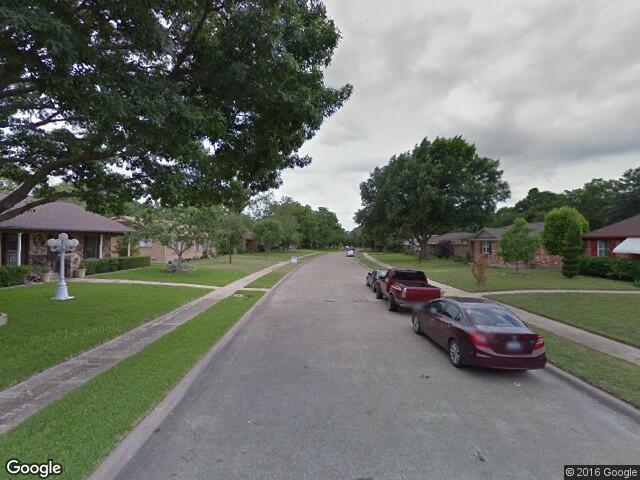}
  \end{subfigure}
  \begin{subfigure}[b]{0.19\textwidth}
    \includegraphics[width=\textwidth]{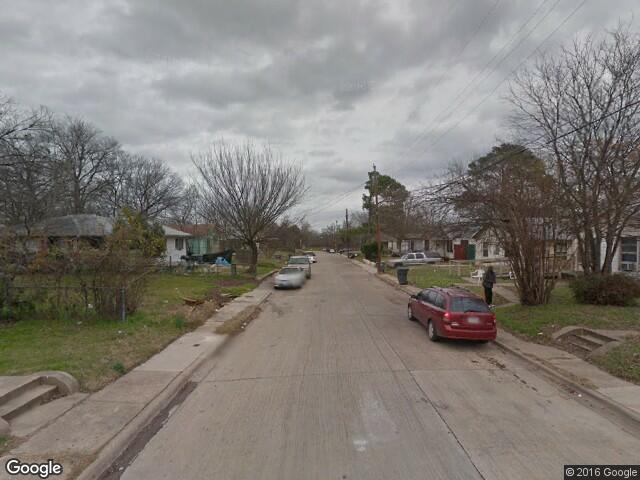}
  \end{subfigure}
  \begin{subfigure}[b]{0.19\textwidth}
    \includegraphics[width=\textwidth]{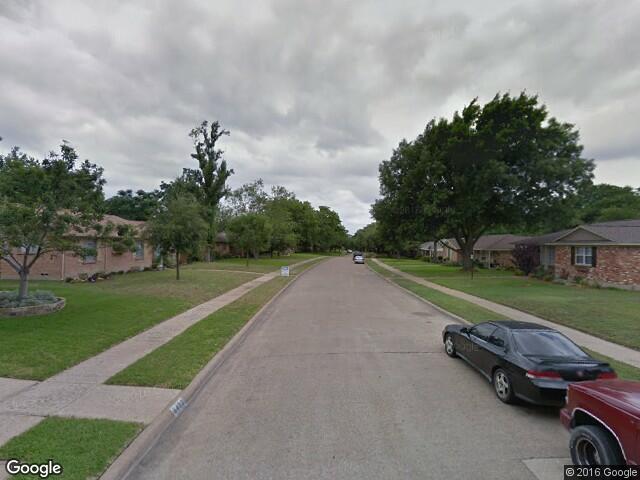}
  \end{subfigure}
  \begin{subfigure}[b]{0.19\textwidth}
    \includegraphics[width=\textwidth]{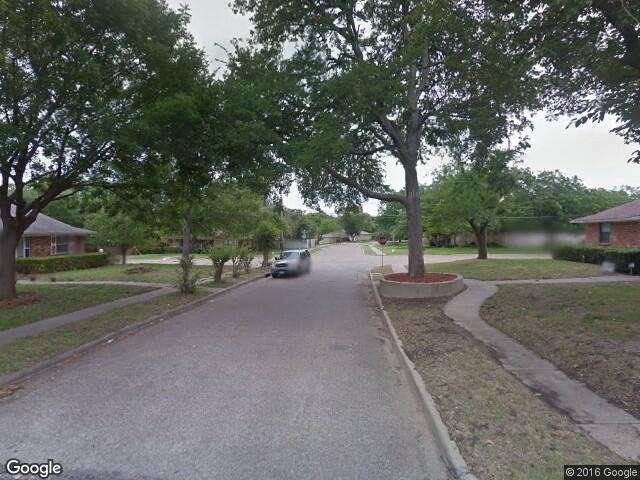}
  \end{subfigure}  
\caption{Rows 1-2: Top 5 high scoring and low scoring (respectively) images predicted by DeepNav-pair navigating to McDonald's in New York (a test city). Rows 3-4: Similar images for navigating to gas station in Dallas (a test city).}
\label{fig:7}
\end{figure*}

If a model learns visual features common to paths leading to destinations, it should pick the correct direction with high confidence near the destination and at major intersections. For a given location, a measure of the confidence of the model for picking one direction is the variance of scores predicted for all directions. We plot this variance at all locations in San Francisco computed from models trained for navigating to Bank of America in Figure~\ref{fig:6}. The figure shows empirically that the DeepNav-pair agent chooses one direction more confidently as it nears a destination, while other models show less of this behavior. Another approach to get an insight into the visual features learnt by the algorithms is to see the which images are most (and least) confidently predicted as pointing to the path to destination. In Figure~\ref{fig:7}, we plot the top- and bottom-5 images (sorted by score) while the DeepNav-pair agent is navigating New York for McDonald's and Dallas for gas station. It can be seen that the CNN correctly learns that center-city commercial areas have a high probability of having a McDonald's establishment, and gas stations are found around intersections of big streets with that have parked cars.\par
\begin{figure*}[pb!]
\centering
  \begin{subfigure}[b]{0.19\textwidth}
    \includegraphics[width=\textwidth]{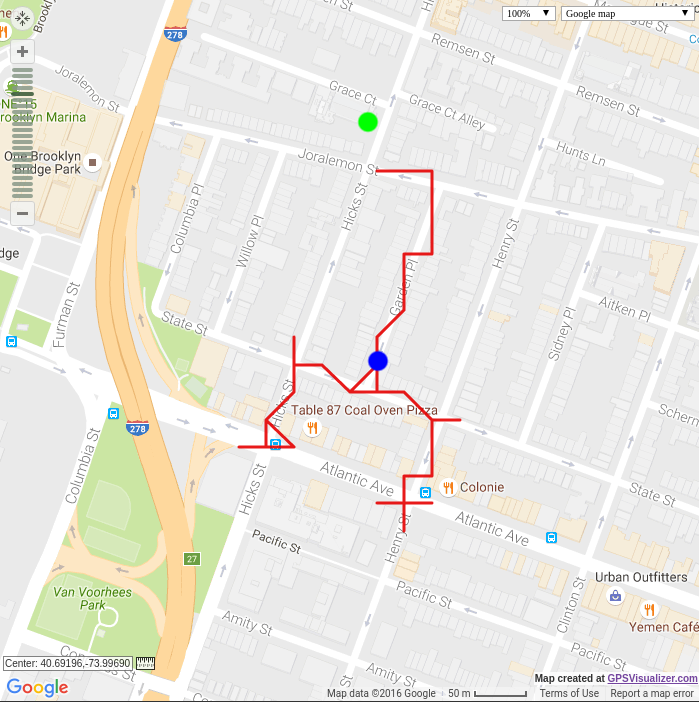}
    \caption*{Found, length = 60\\DeepNav-pair}
  \end{subfigure}
  \begin{subfigure}[b]{0.19\textwidth}
    \includegraphics[width=\textwidth]{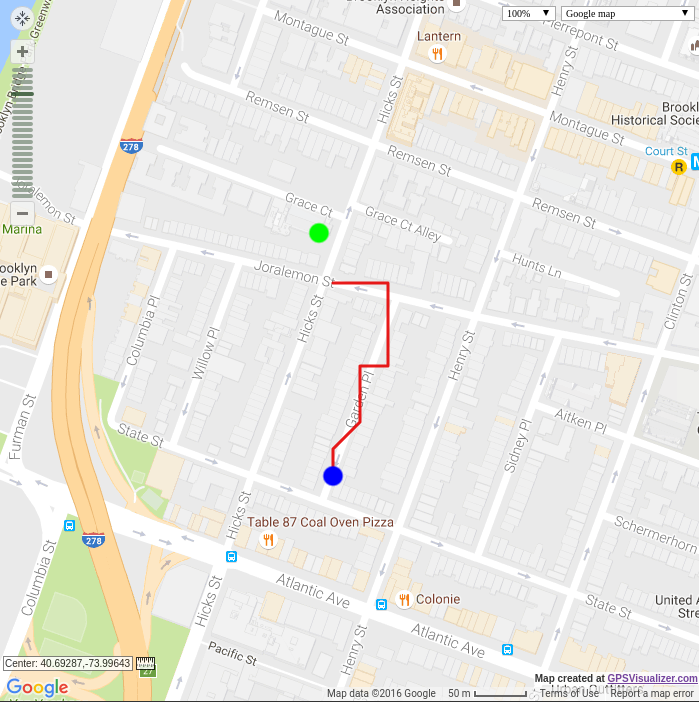}
    \caption*{Found, length = 10\\DeepNav-direction}
  \end{subfigure}
  \begin{subfigure}[b]{0.19\textwidth}
    \includegraphics[width=\textwidth]{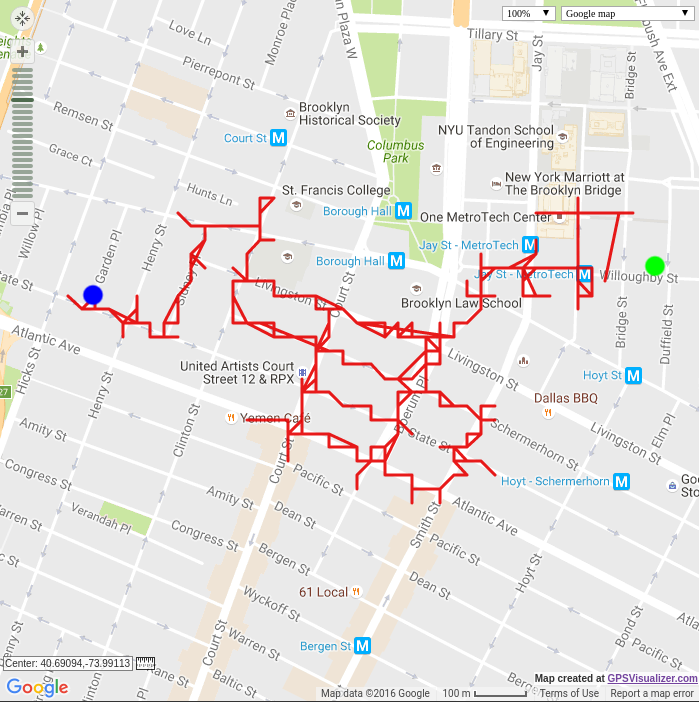}
    \caption*{Found, length = 520\\DeepNav-distance}
  \end{subfigure}
  \begin{subfigure}[b]{0.19\textwidth}
    \includegraphics[width=\textwidth]{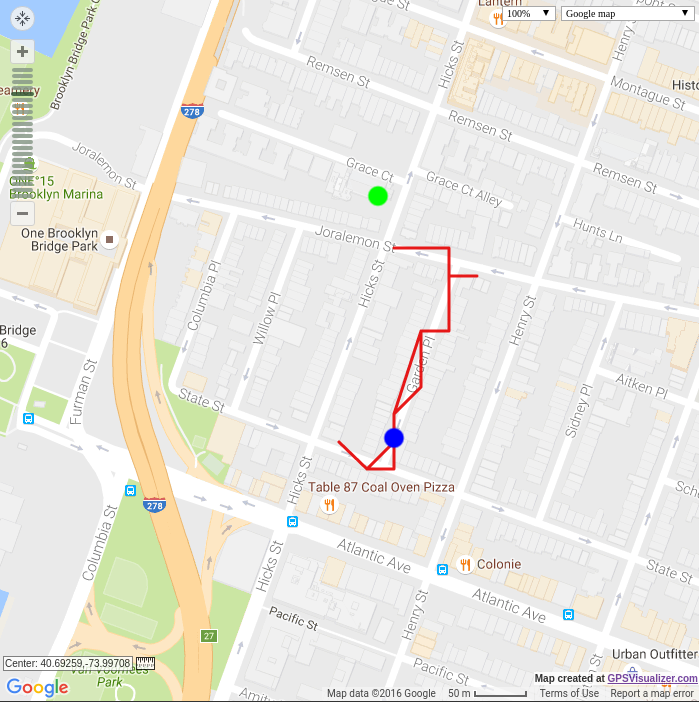}
    \caption*{Found, length = 35\\HOG+SVR~\cite{mcdonalds}}
  \end{subfigure}
  \begin{subfigure}[b]{0.19\textwidth}
    \includegraphics[width=\textwidth]{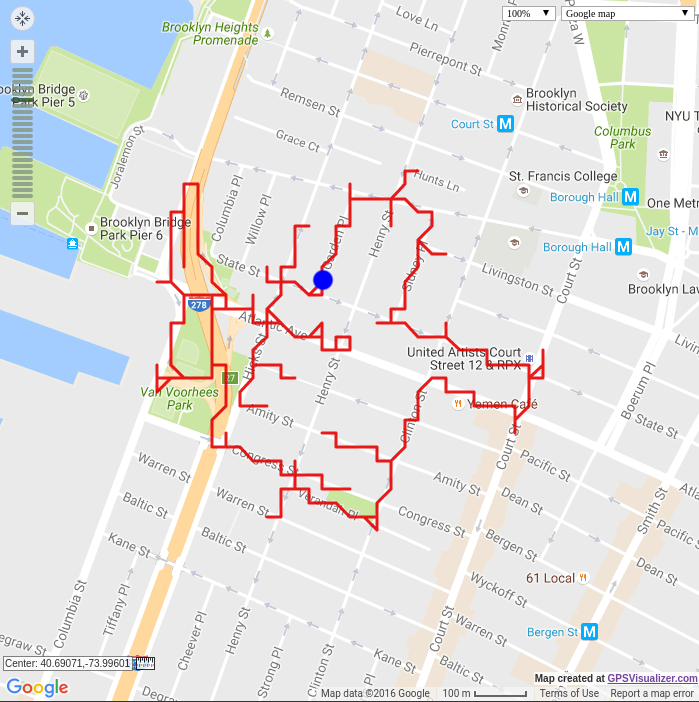}
    \caption*{Not found\\Random walk}
  \end{subfigure}
\caption{Paths for navigating to church in New York (a test city). Blue dot = start, green dot = destination (when found).}
\label{fig:5}
\end{figure*}

Figure~\ref{fig:5} shows some example navigation paths generated by the DeepNav and baseline models while navigating for the nearest church in New York.
\section{Conclusion} \label{sec:6}
We presented 3 convolutional neural network architectures (DeepNav-distance, -direction and -pair) for learning to navigate in large scale real-world environments. These algorithms were trained and evaluated on a dataset of 1 million street-view images collected from 10 large cities. We show how A* search can be used to efficiently generate training labels for images for various DeepNav architectures. We find that data-driven deep convolutional features (DeepNav-distance) outperform a combination of hand-crafted features and SVR~\cite{mcdonalds}. In addition, training the network to `look' in all directions using a Siamese architecture (DeepNav-pair) outperforms networks that are trained to estimate distance to destination (DeepNav-distance) or optimal action (DeepNav-direction).\par
We will release the DeepNav models and training and testing code at \url{https://samarth-robo.github.io/publications.html}.
\FloatBarrier
\begin{table*}[h!]
\resizebox{\textwidth}{!}{
\begin{tabular}{|c|ccccc|ccccc|ccccc|ccccc|c|}
\hline
\multirow{2}{*}{\textbf{Method}} & \multicolumn{5}{c|}{\textbf{Dallas}} & \multicolumn{5}{c|}{\textbf{New York}} & \multicolumn{5}{c|}{\textbf{Phoenix}} & \multicolumn{5}{c|}{\textbf{San Fancisco}} & \multirow{2}{*}{\textbf{Mean}}\\
\cline{2-21}
& \textbf{B} & \textbf{C} & \textbf{G} & \textbf{H} & \textbf{M} & \textbf{B} & \textbf{C} & \textbf{G} & \textbf{H} & \textbf{M} & \textbf{B} & \textbf{C} & \textbf{G} & \textbf{H} & \textbf{M} & \textbf{B} & \textbf{C} & \textbf{G} & \textbf{H} & \textbf{M}\\
\hline
Random walk & 29.24 & 34.28 & 47.43 & 17.73 & 39.40 & 58.90 & 53.93 & 38.57 & 45.00 & 45.37 & 25.63 & 36.23 & 33.73 & 33.98 & 29.33 & 44.47 & 54.04 & 45.88 & 40.00 & 42.21 & 39.77\\
A*	 & 100.00 &	100.00	 & 100.00 &	100.00 & 100.00 & 100.00 &	100.00 &	 100.00 &	100.00	 & 100.00 &	100.00 &	 100.00 & 100.00 & 100.00 & 100.00 &	100.00 & 100.00 & 100.00 &	100.00 & 100.00 & 100.0\\
\hline
HOG+SVR~\cite{mcdonalds} & 48.48 & 56.63 & 62.00 & 06.06 & 42.86 & 76.03 & \textbf{81.55} & 54.29 & \textbf{77.27} & 70.37 & 31.25 & 41.56 & 49.37 & 44.90 & 28.89 & 66.67 & \textbf{70.32} & \textbf{73.40} & 57.32 & 53.49 & 54.63\\
DeepNav-distance	 & 27.27 & \textbf{60.24} & 68.00 & 30.30 & 45.24 & \textbf{80.82} & 66.99 & 48.57 & 68.18 & 69.63 & 37.50 & 54.55 & 49.37 & 46.94 & 35.56 & 63.83 & \textbf{70.32} & 67.55 & \textbf{68.29} & \textbf{65.12} & 56.21\\
DeepNav-direction & 33.33 & 36.14 & 40.67 & 06.06 & 33.33 & 62.33 & 50.49 & \textbf{57.14} & 45.45 & 58.52 & 25.00 & 35.06 & 56.96 & \textbf{55.10} & 31.11 & 45.39 & 61.64 & 60.64 & 42.68 & 13.95 & 42.55\\
DeepNav-pair & \textbf{54.55} & 45.78 & \textbf{78.67} & \textbf{48.48} & \textbf{59.52} & \textbf{80.82} & 67.96	& \textbf{57.14} & 70.45 & \textbf{73.33} & \textbf{43.75} & \textbf{55.84} & \textbf{64.56} & 51.02 & \textbf{48.89} & \textbf{69.50} & 66.21 & 72.87 & 52.44 & 37.21 & \textbf{59.95}\\
\hline
\end{tabular}}
\caption{Success rate of various DeepNav agents compared with random walker, A* and the agent from~\cite{mcdonalds}, $d_s = 470m$. Destination abbreviations: B = Bank of America, C = church, G = gas station, H = high school, M = McDonald's.}
\label{tab:3}
\end{table*}
\begin{table*}[h!]
\resizebox{\textwidth}{!}{
\begin{tabular}{|c|ccccc|ccccc|ccccc|ccccc|c|}
\hline
\multirow{2}{*}{\textbf{Method}} & \multicolumn{5}{c|}{\textbf{Dallas}} & \multicolumn{5}{c|}{\textbf{New York}} & \multicolumn{5}{c|}{\textbf{Phoenix}} & \multicolumn{5}{c|}{\textbf{San Fancisco}} & \multirow{2}{*}{\textbf{Mean}}\\
\cline{2-21}
& \textbf{B} & \textbf{C} & \textbf{G} & \textbf{H} & \textbf{M} & \textbf{B} & \textbf{C} & \textbf{G} & \textbf{H} & \textbf{M} & \textbf{B} & \textbf{C} & \textbf{G} & \textbf{H} & \textbf{M} & \textbf{B} & \textbf{C} & \textbf{G} & \textbf{H} & \textbf{M}\\
\hline
Random walk & 315.11 & 362.67 & 299.07 &	339.28 & 330.13 & 309.59 & 309.04 & 341.49 &	380.03 & 370.48 &	318.77 & 349.53 &	347.89 & 338.43 &	302.59 & 317.05 &	292.74 & 319.62 & 329.60 & 373.98 & 332.35\\
A* & 19.12 & 19.27 & 16.62 & 24.03 &	18.45 &	16.13 &	16.72 &	17.40 &	17.56 &	17.35 &	19.88 &	21.57 &	17.97 &	18.31 &	24.11 &	18.09 &	15.63 &	17.65 &	18.80 &	19.93 &	18.73\\
\hline
HOG+SVR~\cite{mcdonalds} & 244.50 & 326.53 &	257.88 & 164.00 &	\textbf{101.61} & 194.09 &	263.52 & 198.29 & 276.24 &	277.79 & 349.60 &	215.22 & 213.69 & 258.05 &	173.85 & 253.38 &	262.00 & 268.59 &	262.51 & 249.13 & 240.52\\
DeepNav-distance & 321.33 &	289.26 & 295.07 & 163.50 & 303.11 & 322.95 &	247.10 & 220.88	& 284.03 & 267.84 &	270.17 & \textbf{180.29} &	167.69 & 190.26 & 162.94 & 281.57 & 209.27 &	279.06 & 286.59 &	234.18 & 248.85\\
DeepNav-direction & \textbf{43.18} &	\textbf{102.63} 	& \textbf{95.16} &	\textbf{60.00} & 199.50 & \textbf{113.87} &	\textbf{55.19} & \textbf{132.63} & \textbf{127.38} &	\textbf{156.01} & \textbf{132.50} &	189.89 & \textbf{119.71} &	\textbf{166.44} & \textbf{152.71} &	\textbf{68.55} &	\textbf{109.44} & \textbf{128.99} & \textbf{129.29} & \textbf{90.17} &	\textbf{118.66}\\
DeepNav-pair & 414.44 &	246.03 & 243.12 &	433.81 & 158.32 &	223.03 & 239.00 &	246.65 & 331.38 &	269.07 & 273.14 & 209.12 & 174.53 &	231.64 & 35.64 & 222.09 &	263.86 & 256.37 &	226.47 & 247.38 &	257.25\\
\hline
\end{tabular}}
\caption{Average number of steps for successful trials, $d_s = 470m$.}
\label{tab:4}
\end{table*}
\begin{table*}[h!]
\resizebox{\textwidth}{!}{
\begin{tabular}{|c|ccccc|ccccc|ccccc|ccccc|c|}
\hline
\multirow{2}{*}{\textbf{Method}} & \multicolumn{5}{c|}{\textbf{Dallas}} & \multicolumn{5}{c|}{\textbf{New York}} & \multicolumn{5}{c|}{\textbf{Phoenix}} & \multicolumn{5}{c|}{\textbf{San Fancisco}} & \multirow{2}{*}{\textbf{Mean}}\\
\cline{2-21}
& \textbf{B} & \textbf{C} & \textbf{G} & \textbf{H} & \textbf{M} & \textbf{B} & \textbf{C} & \textbf{G} & \textbf{H} & \textbf{M} & \textbf{B} & \textbf{C} & \textbf{G} & \textbf{H} & \textbf{M} & \textbf{B} & \textbf{C} & \textbf{G} & \textbf{H} & \textbf{M}\\
\hline
Random walk & 12.59 &	22.47 &	27.65 &	8.41 & 18.67 & 44.07 & 42.74 & 19.42 &	34.39 &	33.12 &	12.50 &	18.01 &	17.29 &	13.50 &	10.98 &	39.38 &	41.48 &	32.10 &	23.88 & 17.64 & 24.52\\
A*	 & 100.00	& 100.00 & 100.00	& 100.00 & 100.00	& 100.00 & 100.00 & 100.00 & 100.00 & 100.00 & 100.00 & 100.00 & 100.00 & 100.00 & 100.00 & 100.00 & 100.00 & 100.00 & 100.00 & 100.00 & 100.00\\
\hline
HOG+SVR~\cite{mcdonalds} & 25.93 & 44.58 & 46.96	& 18.18	& 26.67	& 66.94 &	\textbf{67.74} &	46.38 &	55.14 & 52.29	& \textbf{25.00}	& 30.88	& 46.99	& 37.14	& 13.73	& 62.50	& 62.68	& 58.52	& 44.90	& 26.42	& 42.98\\
DeepNav-distance	 & \textbf{37.04} & \textbf{56.63} & 53.04 & \textbf{40.91} & 24.44 & \textbf{70.97} & 66.67 & 50.72 & \textbf{60.75} & 55.05 & 40.00	& \textbf{50.00} & 36.14 & \textbf{42.86} & 17.65 & 59.38 &	68.42	& 57.95	& \textbf{47.96}	& \textbf{39.62}	& \textbf{48.81}\\
DeepNav-direction & 25.93 & 27.71	& 34.78	& 4.55 & 15.56 & 54.84 & 39.78 & 53.62 & 50.47 & 52.29 & 20.00 & 25.00 & \textbf{54.22} & 35.71 & 25.49 & 50.00 & 47.37 & 42.05 & 30.61 & 16.98 & 35.35\\
DeepNav-pair & 18.52 & 38.55 & \textbf{69.57} & 31.82	& \textbf{42.22}	& 68.55	& 60.22	& \textbf{56.52}	& 47.66	& \textbf{66.97}	& \textbf{25.00} & 30.88	& 53.01	& 41.43	& \textbf{27.45} & \textbf{65.63} & 63.16 & 56.82	& 43.88	& 30.19 & 46.90\\
\hline
\end{tabular}}
\caption{Success rate of various DeepNav agents compared with random walker, A* and the agent from~\cite{mcdonalds}, $d_s = 690m$.}
\label{tab:5}
\end{table*}
\begin{table*}[h!]
\resizebox{\textwidth}{!}{
\begin{tabular}{|c|ccccc|ccccc|ccccc|ccccc|c|}
\hline
\multirow{2}{*}{\textbf{Method}} & \multicolumn{5}{c|}{\textbf{Dallas}} & \multicolumn{5}{c|}{\textbf{New York}} & \multicolumn{5}{c|}{\textbf{Phoenix}} & \multicolumn{5}{c|}{\textbf{San Fancisco}} & \multirow{2}{*}{\textbf{Mean}}\\
\cline{2-21}
& \textbf{B} & \textbf{C} & \textbf{G} & \textbf{H} & \textbf{M} & \textbf{B} & \textbf{C} & \textbf{G} & \textbf{H} & \textbf{M} & \textbf{B} & \textbf{C} & \textbf{G} & \textbf{H} & \textbf{M} & \textbf{B} & \textbf{C} & \textbf{G} & \textbf{H} & \textbf{M}\\
\hline
Random walk & 458.35 & 408.98 & 408.98 & 450.05 & 488.05 & 344.27 & 338.32 & 508.39 & 429.59 & 405.93 & 496.74 & 472.43 & 486.31 & 471.86 & 523.39 & 298.14 & 373.56 & 399.16 & 406.98 & 463.69 & 431.66\\
A* &	 29.22 & 25.58 & 25.58 & 34.41 & 28.89 & 21.85 & 22.55 & 26.84 & 24.17	& 23.83	& 28.95	& 30.31 & 27.77	& 33.21	& 37.16	& 23.18	& 22.55	& 0.00 & 0.00 & 0.00 & 23.30\\
\hline
HOG+SVR~\cite{mcdonalds} &	465.71 & 417.65	& 349.94 & 283.50 & 280.75 & 279.28 & 293.33 & 323.91 & 325.07 & 282.28 & 282.80 & 261.81 & 334.74 & 265.73 & 254.71 & 249.45 & 336.91 & 320.02 & 323.80 & 374.93 & 315.32\\
DeepNav-distance & 519.80 & 349.06 & 413.03 & 314.67 & 463.91 & 354.48	& 316.48 & 350.34 & 375.23 & 379.45 & 443.00 & 282.21 & 347.23 & 306.93 & 414.11 & 282.84 & 318.82 & 329.16 & 288.30 & 389.62 & 361.93\\
DeepNav-direction &	\textbf{122.71} & \textbf{118.87} & \textbf{104.03} & \textbf{45.00} & \textbf{80.00} & \textbf{147.57} & \textbf{61.51} & \textbf{161.62} & \textbf{142.93} & \textbf{177.65} & \textbf{190.50} & \textbf{238.47} & \textbf{177.31} & \textbf{175.52} & \textbf{178.92} & \textbf{117.06} & \textbf{136.83} & \textbf{120.16} & \textbf{124.83} & \textbf{164.89} & \textbf{139.32}\\
DeepNav-pair &	385.60 & 368.47 & 348.93 & 469.71 & 360.16 & 295.04 & 326.70 & 444.87 & 404.61 & 348.93 & 335.00 & 383.14 & 280.82 & 389.07 & 208.79 & 251.26 & 305.42 & 301.54 & 297.37 & 322.69 & 341.41\\
\hline
\end{tabular}}
\caption{Average number of steps for successful trials, $d_s = 690m$.}
\label{tab:6}
\end{table*}
\begin{table*}[h!]
\resizebox{\textwidth}{!}{
\begin{tabular}{|c|ccccc|ccccc|ccccc|ccccc|c|}
\hline
\multirow{2}{*}{\textbf{Method}} & \multicolumn{5}{c|}{\textbf{Dallas}} & \multicolumn{5}{c|}{\textbf{New York}} & \multicolumn{5}{c|}{\textbf{Phoenix}} & \multicolumn{5}{c|}{\textbf{San Fancisco}} & \multirow{2}{*}{\textbf{Mean}}\\
\cline{2-21}
& \textbf{B} & \textbf{C} & \textbf{G} & \textbf{H} & \textbf{M} & \textbf{B} & \textbf{C} & \textbf{G} & \textbf{H} & \textbf{M} & \textbf{B} & \textbf{C} & \textbf{G} & \textbf{H} & \textbf{M} & \textbf{B} & \textbf{C} & \textbf{G} & \textbf{H} & \textbf{M}\\
\hline
Random walk & 3.20 & 20.13 & 11.22 &	3.28 & 13.89 & 35.32 & 29.73 & 10.86 & 27.29 & 23.54 & 5.59 & 5.00 & 7.30 & 5.36 & 2.79 & 25.71 & 29.75 & 22.21 & 21.69 & 8.19 & 15.60\\
A*	 & 100.00	& 100.00 & 100.00	& 100.00 & 100.00	& 100.00 & 100.00 & 100.00 & 100.00 & 100.00 & 100.00 & 100.00 & 100.00 & 100.00 & 100.00 & 100.00 & 100.00 & 100.00 & 100.00 & 100.00 & 100.00\\
\hline
HOG+SVR~\cite{mcdonalds} & \textbf{20.00} & 26.92 & 41.44 & 6.90 & 20.37 & 55.20 & \textbf{52.05} & 34.48 & \textbf{60.64} & 35.92 & \textbf{23.53} & 9.43 & 38.16 & 26.19 & 11.76 & 56.67 & \textbf{56.28} & \textbf{51.96} & \textbf{40.00} & 21.28 & 34.46\\
DeepNav-distance	 & 16.00 & \textbf{57.69} & \textbf{47.75} & 13.79 &	22.22 &	60.80 &	\textbf{52.05} &	41.38 &	47.87 &	48.54 &	17.65 & 	26.42 &	34.21 &	28.57 &	11.76 &	53.33 &	\textbf{56.28} & 47.49 &	\textbf{40.00} & \textbf{21.28} &	\textbf{37.25}\\
DeepNav-direction & 0.00 & 23.08 &	18.92 &	3.45 & 14.81 & 49.60 & 36.99 & \textbf{46.55} &	32.98 &	40.78 &	17.65 &	18.87 &	43.42 &	\textbf{38.10} &	14.71 &	27.50 &	37.70 &	34.08 &	30.00 &	6.38 & 26.78\\
DeepNav-pair & 4.00 &	34.62 &	46.85 &	\textbf{27.59} &	\textbf{40.74} & \textbf{63.20} &	43.84 &	\textbf{46.55} &	36.17 &	\textbf{50.49} &	17.65 &	\textbf{28.30} & \textbf{48.68} &	33.33 &	\textbf{20.59} & \textbf{69.17} &	47.54 &	29.41 & 35.00 &	17.02 & 37.04\\
\hline
\end{tabular}}
\caption{Success rate of various DeepNav agents compared with random walker, A* and the agent from~\cite{mcdonalds}, $d_s = 970m$.}
\label{tab:7}
\end{table*}
\begin{table*}[h!]
\resizebox{\textwidth}{!}{
\begin{tabular}{|c|ccccc|ccccc|ccccc|ccccc|c|}
\hline
\multirow{2}{*}{\textbf{Method}} & \multicolumn{5}{c|}{\textbf{Dallas}} & \multicolumn{5}{c|}{\textbf{New York}} & \multicolumn{5}{c|}{\textbf{Phoenix}} & \multicolumn{5}{c|}{\textbf{San Fancisco}} & \multirow{2}{*}{\textbf{Mean}}\\
\cline{2-21}
& \textbf{B} & \textbf{C} & \textbf{G} & \textbf{H} & \textbf{M} & \textbf{B} & \textbf{C} & \textbf{G} & \textbf{H} & \textbf{M} & \textbf{B} & \textbf{C} & \textbf{G} & \textbf{H} & \textbf{M} & \textbf{B} & \textbf{C} & \textbf{G} & \textbf{H} & \textbf{M}\\
\hline
Random walk & 311.86 & 326.99 &	473.18 & 341.48 &	373.88 & 311.12 &	346.48 & 507.58 &	365.00 & 377.01 &	397.99 & 474.59 &	576.91 & 528.45 &	371.83 & 366.65 &	389.50 & 371.30 &	292.30 & 423.85 & 396.40\\
A* &	 47.00 & 33.47 & 40.15 & 46.41 &	40.91 &	29.00 &	30.30 &	40.57 &	29.52 &	32.31 &	41.71 &	46.30 &	43.36 &	44.12 &	50.29 &	32.46 &	30.69 &	34.91 &	36.69 &	43.49 &	38.68\\
\hline
HOG+SVR~\cite{mcdonalds} &	\textbf{505.00} & 337.52 &	410.61 & 685.50 &	361.09 & 265.73 &	325.74 & 425.65 &	398.77 & 382.65 &	630.25 & 640.20 & 	438.59 & 403.09 &	450.25 & 367.09 &	413.16 & 362.71 &	380.03 & 422.00 & 430.28\\
DeepNav-distance & 701.75 &	347.78 & 472.38 &	567.00 & 442.25 &	375.66 & 303.05 &	421.50 & 335.11 &	331.08 & 428.00 & 	357.93 & 439.81 &	338.67 & 750.25 & 386.03 & 398.92 &	391.45 & 255.03 & 426.40 & 423.50\\
DeepNav-direction &	- & \textbf{99.67} & \textbf{127.57} &	\textbf{82.00} & \textbf{174.75} & \textbf{189.58} &	\textbf{81.11} & \textbf{206.11} & \textbf{113.00} &	\textbf{155.00} & 350.67 &	437.40 & \textbf{213.03} &	\textbf{222.69} & 367.00 & \textbf{124.39} & \textbf{106.38} &	\textbf{194.48} & \textbf{152.42} &	\textbf{122.33} & \textbf{175.98}\\
DeepNav-pair & 716.00 &	313.74 & 390.29 &	615.25 & 454.91 &	298.76 & 301.97 &	454.26 & 376.53 &	315.00 & \textbf{288.67} &	477.27 & 412.49 &	378.79 & \textbf{316.86} &	331.84 & 361.22 &	351.80 & 292.96 & 264.25 & 387.42\\
\hline
\end{tabular}}
\caption{Average number of steps for successful trials, $d_s =970m$.}
\label{tab:8}
\end{table*}
\FloatBarrier
{\small
\bibliographystyle{templates/ieee}
\bibliography{references}
}

\end{document}